\documentclass{article}

\usepackage{microtype}
\usepackage{graphicx}
\usepackage{booktabs} % for professional tables
\usepackage{amsmath,amssymb,amsfonts}
\usepackage{amsthm}
\usepackage{booktabs}
\usepackage{fixmath}
\usepackage{textcomp}
\usepackage{mathtools}
\usepackage[dvipsnames]{xcolor}
\usepackage[export]{adjustbox}
\usepackage{float}
\usepackage{multirow}
\usepackage[hidelinks]{hyperref}
\usepackage[flushleft]{threeparttable}

\usepackage{wrapfig}
\usepackage{caption}
\usepackage{graphicx,subcaption,lipsum}
\usepackage{upgreek}
\usepackage{pifont}
\usepackage{bm}
\usepackage[flushleft]{threeparttable} % 
\usepackage[symbol]{footmisc}
\usepackage{dblfloatfix}
\usepackage{hyperref}

\usepackage[accepted]{icml2024}

\usepackage{amsmath}
\DeclareMathOperator*{\argmax}{argmax} 
\DeclareMathOperator*{\argmin}{argmin} 

\usepackage{amssymb}
\usepackage{mathtools}
\usepackage{amsthm}

\DeclarePairedDelimiterX{\infdivx}[2]{(}{)}{%
  #1\;\delimsize\|\;#2%
}
\newcommand{\infdiv}{D_{KL}\infdivx}

\usepackage[capitalize,noabbrev]{cleveref}

\theoremstyle{plain}
\newtheorem{theorem}{Theorem}[section]
\newtheorem{proposition}[theorem]{Proposition}

\theoremstyle{definition}

\theoremstyle{remark}

\usepackage[textsize=tiny]{todonotes}

\icmltitlerunning{An Unsupervised Approach for Periodic Source Detection in Time Series}

\begin{document}

\twocolumn[
\icmltitle{An Unsupervised Approach for Periodic Source Detection in Time Series \\}



\icmlsetsymbol{equal}{*}

\begin{icmlauthorlist}
\icmlauthor{Berken Utku Demirel}{y}
\icmlauthor{Christian Holz}{y}
\end{icmlauthorlist}

\icmlaffiliation{y}{Department of Computer Science, ETH Zurich, Switzerland}

\icmlcorrespondingauthor{Berken Utku Demirel}{berken.demirel@inf.ethz.ch}

\icmlkeywords{Machine Learning, ICML}

\vskip 0.3in
]



\printAffiliationsAndNotice{}  % leave blank if no need to mention equal contribution

\begin{abstract}
Detection of periodic patterns of interest within noisy time series data plays a critical role in various tasks, spanning from health monitoring to behavior analysis.
Existing learning techniques often rely on labels or clean versions of signals for detecting the periodicity, and those employing self-supervised learning methods are required to apply proper augmentations, which is already challenging for time series and can result in collapse---all representations collapse to a single point due to strong augmentations.
In this work, we propose a novel method to detect the periodicity in time series without the need for any labels or requiring tailored positive or negative data generation mechanisms with specific augmentations.
We mitigate the collapse issue by ensuring the learned representations retain information from the original samples without imposing any random variance constraints on the batch.
Our experiments in three time series tasks against state-of-the-art learning methods show that the proposed approach consistently outperforms prior works, achieving performance improvements of more than 45--50\%, showing its effectiveness.

\small{\texttt{Code:~\href{https://github.com/eth-siplab/Unsupervised_Periodicity_Detection}{https://github.com/eth-siplab/\\Unsupervised\,\_\,Periodicity\,\_\,Detection}}}
\end{abstract}

\section{Introduction}
\label{sec:introduction}
Periodic structure in time series data holds significant importance in monitoring individuals' behaviors and health~\cite{sandvik_heart_1995,nurses_and_resp}. 
A wide range of tasks related to time series, including step counting~\cite{step_dead} and heart rate monitoring~\cite{JAMA_hrv}, can reveal crucial information about one's health situation.
% encompassing a wide range of applications, from step counting~\cite{step_dead} to heart rate monitoring~\cite{JAMA_hrv}.
Moreover, on a daily basis, people generate substantial amounts of time series data from smartphones or wearable devices~\cite{nature_wearable}, which can be used to provide insights into their behaviors and health~\cite{JMIR,wearables_cvd}.
However, labeling these huge amounts of temporal data is challenging, expensive, and resource-intensive. 
For instance, if the desired task is to monitor the respiration or heart rate, collecting gold-standard synchronized signals for obtaining ground truth information from a medical sensor or video is not only time-consuming but also expensive~\cite{simper, CLOCS}. 
Moreover, obtaining ground truth is not feasible in daily life, i.e., non-controlled environments outside of the lab, due to the potential discomfort it may impose on individuals~\cite{gao_wearable_2023}, which can be caused by privacy concerns, such as the use of video recording~\cite{privacy_video}, or the use of bulky electrodes with gels, as in the case for medical sensors~\cite{bulky_ecg}.

The self-supervised learning (SSL) paradigm provides a solution to overcome this problem by exploiting unlabelled data to formulate pretext tasks such as predicting the rotation of images~\cite{rotation_prediction}, or contexts~\cite{Context_prediction} to learn the invariant representations of samples~\cite{understand_collapse} to the applied transformation.
The effectiveness of representations on downstream tasks directly depends on the inter-sample semantic similarity relations~\cite{balestriero2022contrastive, chaos} that are created through tailored data augmentation techniques to preserve semantics depending on the task~\cite{InfoMin, SimCLR} and input characteristics~\cite{demirel2023chaos}.
% Moreover, recently, it was shown that the role of data augmentation in SSL is to create new samples between different intra-class samples such that they become more alike~\cite{chaos}.
However, even when the downstream task is known, creating samples that keep the task-relevant information intact is more challenging in noisy quasi-periodic time series due to the complexity of the dynamical data generation mechanisms~\cite{complex_mechanism}, where strong augmentations can cause model collapse~\cite{understand_collapse} while weak augmentations can hinder model convergence~\cite{InfoMin}.

Considering these limitations, in this work, we propose a novel method that regularizes the parametric learners to detect periodic components in time series data by punishing the randomness in a sequence while capturing as much information as possible about the periodicity without requiring any specialized data augmentation technique.
Since our proposed method bypasses the intricate data augmentation process required for generating positive or negative samples to learn the relations or invariant representations between samples, we prevent the model collapse issue by relating the learned representations with original samples from which they originated, rather than decorrelating the representations with each other or imposing random variance constraints on the batch.  
Our approach argues that every learned representation should retain a remnant of the sample from which it originates.
The contributions can be summarized as follows:

\begin{itemize}
\item We propose a novel set of regularizers that consider the generation mechanism of time series data to detect periodic patterns of interest in the presence of noise.

\item We introduce a new approach to prevent collapse in unsupervised periodicity detection by relating learned representations with samples from which they originated, resulting in relaxed assumptions for batch statistics.

\item We demonstrate that the proposed regularizers help the learners extract useful representations without requiring any supervision or specialized time series augmentations for generating positive/negative samples.

\end{itemize}

% The remainder of this paper is organized as follows.
% Section~\ref{sec:Prop_method} describes the proposed method. 
% Section~\ref{sec:Exp_setup} describes the experimental setup.
% Section~\ref{sec:Results} discusses the results. 
% Finally, we lay out our conclusion in Section~\ref{sec:Conclusion}.

\section {Method}
\label{sec:Prop_method}

\subsection{Notations}
We use lowercase symbols $(x)$ to denote scalar quantities, bold lowercase symbols $(\boldsymbol{\mathrm{x}})$ for vectors, e.g., time series, and capital letters $(X)$ for random variables. 
The parametric functions is represented as $f_{\theta}(.)$ where $\theta$ is the parameters.
The discrete Fourier transformation is denoted as $\mathcal{F}(.)$, yielding a complex variable as $\mathcal{F}(\boldsymbol{\mathbf{x}}) \in \mathbb{C}$.
The detailed calculations for each operation are given in the Appendix~\ref{appen:notations}.

\subsection{Objective}
% The objective is formulated as follows.
Given a dataset $\mathcal{D} = \{  (\boldsymbol{\mathrm{x}}_i) \}_{i=1}^K$ where each $\boldsymbol{\mathrm{x}}_i$ consists of real-valued sequences with length L, $(x_1, x_2, \dots, x_L)$, which are sampled uniformly.
The objective is to train a learner $f_{\theta}: X \rightarrow \mathbb{R}^l$ which seeks to learn periodic representations of interest in the data such that when it is evaluated on another set $\mathcal{D}_l = \{  (\boldsymbol{\mathrm{x}}_i, y_i) \}_{i=1}^M$, the representations $\mathbf{z}_i=f_{\theta}(\mathbf{x}_i)$ will mainly contain desired periodic source, which can be detected i.e., $\argmax\limits_k |\mathcal{F}(\mathbf{z}_i)_k|  =  {y_i} \in \mathbb{R}^+$.

Since our objective is to detect the periodic source of interest in the uniformly sampled time series data without using any labels, we proposed a set of regularizers designed to help the learners extract useful representations that capture the desired periodicity.
In the following sections, we describe the proposed regularizers and outline their primary objectives.
Our first proposed regularizer decreases the randomness of the overall time series sequence while maximizing the periodicity by minimizing the entropy of the spectra as follows. 

\begin{figure}[b]
    \centering
    \includegraphics{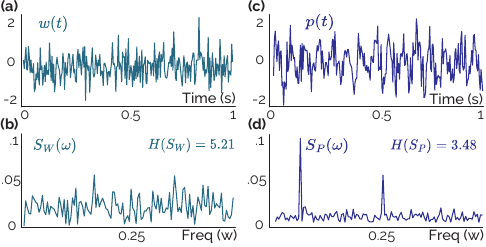}
    \caption{\textbf{(a)} A time sequence sample $w(t)$ from a Gaussian process \textbf{(b)} with its normalized spectra $S_W(\omega)$, \textbf{(c)} a periodic sequence in the presence of added random noise, denoted as $p(t)$ \textbf{(d)} the corresponding spectra of the sequence $p(t)$, denoted as $S_P(\omega)$.}
    \label{fig:spectral_entropy}
\end{figure}

\subsubsection{Maximizing the Periodicity}
\begin{proposition}[Maximizing periodicity]\label{prop:spectral_entropy}
Minimizing the spectral entropy of sequential data is nothing but maximizing its periodicity in the time domain.
\begin{equation*}\label{eq:spectral_entropy_prop}
\begin{gathered}
\theta^* = \argmin_{\theta} \mathcal{L}_{se} = \argmax_{\theta} \mathcal{I}(\mathbf{y}; \mathbf{x}), 
\\
\text{where}  \hspace{2mm} \mathcal{L}_{se} = - \sum_{w}^{} S_{f_{\theta}}(\omega) \log S_{f_{\theta}}(\omega),
\\
S_{f_{\theta}}(\omega) = \frac{S_{f_{\theta}}(\omega)}{\sum_{\omega} S_{f_{\theta}}(\omega)}, \hspace{1mm} S_{f_{\theta}}(\omega) = \left| \sum_{n}^{} f_{\theta}(\mathbf{x})_n e^{-j\omega n}  \right|^2 
\end{gathered}
\end{equation*}
\end{proposition}
\begin{proof}
The spectral entropy is maximum (periodicity is minimum) when the sequence generation process is a Gaussian with zero mean, $\sigma^2$ variance, i.e., $W(t) \sim \mathcal{N}(0, \sigma^2_w)$.
\begin{gather}
    \mathbb{E}[W(t)] =0, \hspace{2mm} R_{WW}(\tau) = \mathbb{E}[W(t+\tau) W^*(t)]
    \\
    R_{WW}(\tau) = \sigma^2 \delta(\tau) \xrightarrow[]{\mathcal{F(.)}} S_{W}(\omega) = \sigma_{\omega}^2
    \\
    H(S_{W}) \geq H(S_{X}), \forall \mathbf{x} \in X 
\end{gather}
The spectral entropy is minimum if the sequence generation process is periodic, i.e., $R_{PP}(\tau) = R_{PP}(\tau + T)$.
\begin{gather}
    R_{PP}(\tau) = \sum_{n=-\infty}^{\infty} \alpha_n e^{j \omega_0 n \tau}, \hspace{2mm} \omega_0 = \frac{2 \pi}{T}
    \\
    \alpha_n = \frac{1}{T} \int_{-T/2}^{T/2} R_{PP}(\tau) e^{-j \omega_0 n \tau} d\tau
    \\
    S_P(w) = 2\pi\sum_{n-\infty}^{\infty} \alpha_n \delta(\omega-n\omega_0)
    \\
    H(S_{P}) \leq H(S_{X}), \forall \mathbf{x} \in X  
\end{gather}
Therefore, $H(S_{P}) \leq \mathcal{L}_{se} \leq H(S_{W})$.
\end{proof}
The last inequality completes the proof by showing that minimizing the spectral entropy of sequential data is equivalent to maximizing its periodicity.
A detailed proof, with the bandlimited discrete case, can be found in the Appendix~\ref{appen:proof}.
We also provide an intuitive demonstration of samples from data generation processes and corresponding spectra in Figure~\ref{fig:spectral_entropy} where it can be observed that the periodic data has lower entropy compared to the sample from Gaussian distribution.

Although this regularizer can induce periodicity in sequences, it will cause the model to collapse to a single point and produce a constant output.
The existing solutions for preventing collapse generally incorporate the notion of class balance across the dataset~\cite{krause_nips} or batch~\cite{vicreg,SiNC}.
Such approaches, however, are overly optimistic since during SSL training, where there is no label information, this regularizer may punish the models even for learning correct representations.
Alternatively, the models can learn task-unrelated information to introduce diversity in representations.
Therefore, to prevent the degenerate solution, we have incorporated a secondary regularizer that ensures the extracted representations retain information from the original samples.
\subsubsection{Preventing the Collapse}
\begin{proposition}[Degenerate solution]\label{prop:KL_divergence}
The relative entropy between the spectral distributions of samples and extracted representations is upper-bounded by a degenerate solution.
\begin{equation}\label{eq:KL_divergence}
\begin{gathered}
\infdiv{\mathbb{P}}{\mathbb{Q}} > \mathcal{L}_{ds} \geq 0, \hspace{2mm} \text{where}  \hspace{2mm} \text{Var}(\mathbb{Q}) = 0,
\\
\text{and}  \hspace{2mm} \mathcal{L}_{ds} = \sum_{\omega} S_X(\omega) \log \frac{S_X(\omega)}{S_{f_{\theta}}(\omega)},
\\
S_X(\omega) = \frac{S_X(\omega)}{\sum_{\omega} S_X(\omega)}, \hspace{1mm} S_X(\omega) = \left| \sum_{n}^{} x_n e^{-j \omega n}  \right|^2 
\end{gathered}
\end{equation}
\end{proposition}
\begin{proof}
There exists a sample with a spectrum $S_X(\omega)$ for which a collapsed learner, trained to maximize periodicity (minimize spectral entropy), results in an infinite divergence of spectral distributions between samples and embeddings.
\begin{gather}
    \argmin_{\theta} \mathcal{L}_{se} \implies \lim_{\mathcal{L}_{se}\to 0} S_{f_{\theta}}(\omega) \to 0, \hspace{1mm} \exists \omega \in \upOmega
    \\
    \exists \omega \in (-\infty, \infty), \quad S_{f_{\theta}}(\omega) = 0 \wedge S_X(\omega) \neq 0,
\end{gather}
yields $D_{KL}(\mathbb{P} \hspace{1mm} \lVert \hspace{1mm} \mathbb{Q})\hspace{-0.5mm}\rightarrow \hspace{-0.5mm} \infty$, \hspace{1mm} $S_X(\omega) \sim \mathbb{P}$, $S_{f_{\theta}}(\omega) \sim \mathbb{Q}$ 
\end{proof}

We provide an intuitive illustration of the proposition with a sample in Figure~\ref{fig:spectral_divergence} and the detailed proof in Appendix~\ref{appen:proof}.

The proposition~\ref{prop:KL_divergence} suggests that degenerate solutions can be prevented if the learner seeks to minimize the relative entropy between the spectra of samples and extracted representations.
In other words, this proposed regularization technique helps the learner to generate diverse representations according to the input distribution, preventing the collapse where inputs are mapped to the same vector, while avoiding the imposition of random variances within a batch.

The usage of these two regularizers together also eliminates the requirement for generating positive or negative samples to construct embedding space with specialized augmentations which is known to be particularly challenging for non-stationary time series data.
Moreover, most of the unsupervised learning methods prevent model collapse by repelling the negative pairs which are randomly chosen from the batch~\cite{SimCLR}, selected from memory banks according to the learned relevance~\cite{He2019MomentumCF}, or enforcing distinct embeddings within a batch~\cite{vicreg}.
However, since there is no label information during training, these methods can repel/decorrelate representations of samples from the same category which can lead to suboptimal embedding space.
Therefore, to mitigate this problem, we take a different approach and govern the representation space through two opposing regularizers, seeking to find the periodicity of interest in the noisy time sequence data without causing significant distortions to the features. 
\begin{figure}[b]
    \centering
    \includegraphics{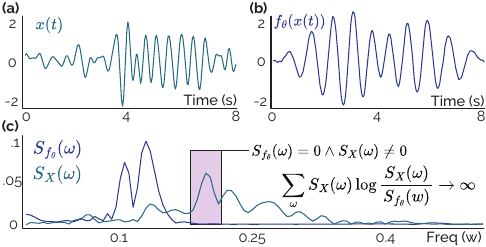}
    \caption{\textbf{(a)} 8-second time sequence $x(t)$, \textbf{(b)} its obtained representation from a model that is trained to maximize periodicity of the signal, \textbf{(c)} the normalized spectra of the sample and representation.
    As the model is trained to minimize the spectral entropy, the values of $S_{f_{\theta}}(\omega)$ diminish to zero, making the divergence infinity.}
    \label{fig:spectral_divergence}
\end{figure}

When we calculate these two loss functions, we specify a bandlimited frequency range of interest $f^*$ instead of considering the entire continuous spectrum while minimizing the overall power in frequencies outside of this defined range $f^{'}$.
Minimizing frequencies outside of the specified range, essentially applying a bandpass filter, is a common approach in the literature~\cite{SiNC, the_way_to_my_heart}. 
We observe marginal improvements with this addition while the improvement is diminished with better-designed filters.
The overall loss for training models is given in Equation~\ref{eq:overall_loss}.
\begin{multline}\label{eq:overall_loss}
    \mathcal{L} =  - \sum_{f^*}^{} S(\omega) \log S(\omega) + \sum_{f^*}^{} S_X(\omega) \log \frac{S_X(\omega)}{S(\omega)}
    \\
    + \sum_{f^{'}}^{} S(\omega), \hspace{2mm} \omega = 2 \pi f \hspace{1mm} \text{and} \hspace{1mm} f^{'} = \mathbb{U} \setminus f^*
\end{multline}
The learner which is regularized by the proposed loss function is similar to an adaptive sinusoidal selector based on the Fourier transform. 
% The underlying concept of this learner is rooted in the idea that the proposed regularizers as adaptive sinusoidal selectors based on the Fourier transform.
In uniformly sampled time sequences, periodicity is typically identified by the highest value in the Fourier transform~\cite{periodicity_fft}. 
However, this approach fails when there exist multiple periodicities within the frequency range of interest, or when the desired periodic pattern is obscured by noise. 
The proposed training regularizers help the model learn the desired pattern in the time domain such that it can detect the signal even under the noise.
The characteristic of this loss is provided by the minimization of the spectral entropy. 
This is evident when considering that, if the spectral entropy term is excluded from the overall loss function, the optimization converges to a bandpass-filtered Fourier transform of the original signal.

\section{Experiments}
\label{sec:Exp_setup}	
We compare our method with several techniques, from traditional heuristic-based methods such as autocorrelation and Fourier transform-based approaches, as well as learning-based methods including supervised and self-supervised.
% supervised, self-supervised, and label-free approaches, using data from real life with different noise levels to provide a more comprehensive evaluation.
% Complete training details and hyper-parameter settings for datasets and baselines are provided in Appendix~\ref{appendix:implementation_details}.

% -------------------------------------------------- %
\subsection{Datasets}
\label{ref:datasets}
We conducted experiments on eight datasets across three tasks, including heart rate (HR) estimation from electrocardiogram (ECG) and photoplethysmography (PPG) signals, step counting using inertial measurements (IMUs), and respiratory rate (breathing) estimation from PPG signals.
We provided short descriptions of each dataset below, and further detailed information can be found in Appendix~\ref{appendix:datasets}.

\paragraph{Heart rate} 
For ECG, we used PTB-XL~\cite{ptb} to investigate performance in a dataset with cardiovascular diseases, WESAD~\cite{WESAD}, and DaLiA~\cite{DeepPPG} to study the real-life settings, where the signals are corrupted with different level of noise as they are collected during activities like cycling.
We used the IEEE SPC12 with 22~\cite{TROIKA}, and DaLiA~\cite{DeepPPG} for PPG-based HR prediction.

\paragraph{Respiratory rate} 
We used CapnoBase~\cite{capno} with two different window sizes and BIDMC~\cite{bidmc} for estimating respiratory rate (RR) from the PPG, where both datasets provide true RR using gold standard respiratory signals and capnography.

\paragraph{Step counting} 
We used the Clemson dataset~\cite{clemson}, which is proposed to improve pedometer evaluation. 
We conducted experiments using inertial measurements from the wrist with both regular and semi-regular walking settings where labels are available through videos.

\subsection{Baselines}
\label{ref:baseline}

\paragraph{Self-supervised} 
We compared our method with self-supervised learning techniques within the linear evaluation setting~\cite{SimCLR}, including SimCLR~\cite{SimCLR}, NNCLR~\cite{NNCLR}, BYOL~\cite{BYOL}, TS-TCC~\cite{tstcc}, TS2Vec~\cite{TS2VEC}, VICReg~\cite{vicreg}, Barlow Twins~\cite{barlow_twins}, and SimPer~\cite{simper}.

\paragraph{No supervision} 
We conducted experiments with a learning-based method that requires no labels for fine-tuning. 
Specifically, we compared our work with SiNC~\cite{SiNC} which is a non-contrastive unsupervised learning framework for periodicity detection without using any labels while imposing random variance constraints on the batch.

\paragraph{Supervised} 
We included two supervised models: a fully convolutional network and a convolutional network with LSTM~\cite{LSTM_paper}, both commonly used in time series tasks previously~\cite{KDD_paper}.

\paragraph{Traditional}
We also included traditional heuristic-based methods, i.e., the Fourier transformation and autocorrelation function, where both are widely used before~\cite{saul_periodic} to detect periodicity in time series.
Additionally, we compared our method with RobustPeriod~\cite{robustperiod} which is a rule-based periodicity detection approach by integrating information from both time and frequency domains.
More details for each baseline are given in Appendix~\ref{appendix:baselines}.

\subsection{Implementation}
\label{ref:implementation}

\paragraph{Self-supervised}
We follow the same implementation setup with previous works~\cite{KDD_paper, demirel2023chaos} for self-supervised learning in time series.
Specifically, we use a combination of convolutional with LSTM-based network, which shows superior performance in many time series tasks~\cite{KDD_paper, CorNET}, as backbones for the encoder $f_{\theta}(.)$ where the projector is two fully-connected layers. 
During pre-training, we use InfoNCE (for contrastive learning-based methods) as the loss function, which is optimized using Adam~\cite{Adam} with a learning rate of $0.003$.
We train the models with a batch size of 256 for 120 epochs and decay the learning rate using the cosine decay.
After pre-training, we train a single linear layer classifier on features extracted from the frozen pre-trained network, i.e., linear probing.
Although we aim to enhance performance by experimenting with various architectures and hyperparameters, we consistently observed the best average performance across datasets with the specified implementation details, similar to previous studies. 
Detailed hyperparameters and augmentations for each SSL technique can be found in Appendix~\ref{appendix:baselines_SSL}.
% ~\footnote{The source code is available at \href{https://github.com/eth-siplab/Unsupervised_Periodicity_Detection}{https://github.com/eth-siplab/Unsupervised\_Periodicity\_Detection}}
\begin{table*}[hb!]
\centering
\caption{Performance comparison of ours with other methods in \textit{ECG} datasets for HR detection}
\begin{adjustbox}{width=2\columnwidth,center}
\label{tab:performance_ecg}
\renewcommand{\arraystretch}{0.7}
\begin{tabular}{@{}lllllllllll@{}}
\toprule
\multirow{2}{*}{Method} & \multicolumn{3}{l}{PTB-XL} & \multicolumn{3}{l}{DaLiA$_{ECG}$} & \multicolumn{3}{l}{WESAD} \\ 
\cmidrule(r{15pt}){2-4}  \cmidrule(r{15pt}){5-7}  \cmidrule(r{15pt}){8-10} \\ 
& MAE$\downarrow$ & RMSE$\downarrow$ & $\rho$$\uparrow$ & MAE$\downarrow$ & RMSE$\downarrow$ & $\rho$$\uparrow$ & MAE$\downarrow$ & RMSE$\downarrow$ & $\rho$$\uparrow$ \\
\midrule
\textit{Heuristic} & & & & & & & & & \\
Fourier & 10.51 & 28.65 & 49.12 & 4.06 & 15.46 & 73.73 & 4.88 & 18.57 & 56.69\\
Autocorrelation & 8.64 & 19.93 & 59.63 & 7.07 & 14.07 & 76.34 & 4.19 & 8.046 & 89.90 & \\
RobustPeriod & 72.79 & 80.87 & -2.93 & 12.79 & 23.21 & 37.49 & 25.16 & 30.18 & 10.23 & \\
\midrule
\textit{Supervised} & & & & & & & & & \\
DCL & 6.08\small$\pm$0.78 & 14.23\small$\pm$ 1.43 & 76.40\small$\pm$6.02 & 3.91\small$\pm$0.37 & 6.96\small$\pm$1.06 & 91.42\small$\pm$2.5 & 9.14\small$\pm$0.78 & 7.23\small$\pm$0.75 & 85.44\small$\pm$6.32 \\
CNN & 9.09\small$\pm$0.27 & 16.37\small$\pm$0.33  & 66.11\small$\pm$1.75 & 5.39$\pm$0.16 & 8.36\small$\pm$0.15 & 89.44\small$\pm$0.72 & 10.69\small$\pm$0.67 & 13.57\small$\pm$1.01 & 50.10\small$\pm$12.5 \\
\midrule
\textit{Self-Supervised} & & & & & & & & & \\
SimCLR & 9.13\small$\pm$0.73 & 18.78\small$\pm$0.12 & 70.67\small$\pm$0.55 & 6.23\small$\pm$1.10 & 14.38\small$\pm$1.67 & 68.46\small$\pm$5.25 & 7.09\small$\pm$0.24  & 11.05\small$\pm$0.51 & 65.80\small$\pm$2.96 \\
NNCLR & 10.05\small$\pm$2.06 & 18.74\small$\pm$1.96 & 61.50\small$\pm$7.71 & 8.33\small$\pm$1.85 & 10.89\small$\pm$2.08 & 79.06\small$\pm$8.01 & 8.84\small$\pm$1.14 & 11.64\small$\pm$1.33 & 59.48\small$\pm$9.39 \\
BYOL & 11.92\small$\pm$3.44 & 24.07\small$\pm$8.94 & 50.67\small$\pm$19.7 & 7.39\small$\pm$0.27 & 10.55\small$\pm$0.39 & 81.57\small$\pm$0.93 & 12.21\small$\pm$0.35 & 15.93\small$\pm$0.45 & 28.99\small$\pm$2.55 \\
TS-TCC & 10.13\small$\pm$0.52 & 18.79\small$\pm$0.65 & 54.23\small$\pm$4.62 & 5.13\small$\pm$0.08 & 7.75\small$\pm$0.13 &90.65\small$\pm$0.09 & 6.16\small$\pm$0.29 & 8.27\small$\pm$0.40 & 83.30\small$\pm$1.68 
\\
TS2Vec & 9.52\small$\pm$3.54 & 24.26\small$\pm$6.46 & 52.98\small$\pm$10.08 & 4.78\small$\pm$0.23 & 7.14\small$\pm$0.18 & 88.95\small$\pm$0.13 & 6.24\small$\pm$0.17 & 7.33\small$\pm$0.37 & 84.12\small$\pm$2.10 
\\
VICReg & 15.09\small$\pm$0.23 & 23.51\small$\pm$0.34 & 01.35\small$\pm$0.53 & 13.12\small$\pm$4.66 & 18.82\small$\pm$4.30 &66.18\small$\pm$10.15 & 10.29\small$\pm$0.62 & 13.93\small$\pm$0.84 & 45.27\small$\pm$5.33  \\
Barlow Twins & 13.78\small$\pm$3.73 & 22.96\small$\pm$5.30 & 43.90\small$\pm$10.03 & 13.47\small$\pm$0.50 & 18.02\small$\pm$0.58 & 56.97\small$\pm$3.23 & 11.61\small$\pm$0.34 & 15.09\small$\pm$0.50 & 31.19\small$\pm$3.37  \\
SimPer & --- & --- & --- & 10.11\small$\pm$2.53 & 16.12\small$\pm$1.57 & 60.62\small$\pm$7.23 & 8.16\small$\pm$0.27 & 12.13\small$\pm$0.62 & 52.17\small$\pm$3.21  \\
\midrule
\textit{No Supervision} & & & & & & & & & \\
SiNC & 3.97\small$\pm$0.01 & 13.71\small$\pm$0.03 & 79.27\small$\pm$0.13 & 1.72\small$\pm$0.013 &  2.65\small$\pm$0.10 & 98.96\small$\pm$0.03 & 2.41\small$\pm$0.01 & 4.13\small$\pm$0.13 & 93.21\small$\pm$0.14 \\
Ours & \textbf{3.75}\small$\pm$0.02 & \textbf{13.58}\small$\pm$0.03 & \textbf{79.30}\small$\pm$0.17 & \textbf{1.29}\small$\pm$0.001 & \textbf{2.08}\small$\pm$0.01 &\textbf{99.12}\small$\pm$0.01& \textbf{2.13}\small$\pm$0.01 & \textbf{3.88}\small$\pm$0.06 & \textbf{93.52}\small$\pm$0.61 & \\
\bottomrule
\end{tabular}
\end{adjustbox}
\end{table*}

\paragraph{No supervision} 
Unlike the self-supervised methods, we employ the backbone as a representation learner without having a non-linear projection layer.
In our initial experiments, we observed that LSTM or temporal dilations cause loss of frequency information similar to~\cite{SiNC}, we, therefore, use U-Net~\cite{U_net} as the encoder architecture with single dimensional layers for time series.
We use the proposed loss function in Equation~\ref{eq:overall_loss}, which is optimized using Adam with a learning rate of $0.001$.
We train the model with a batch size of 512 and reduce the learning rate by half when the training loss stops decreasing for 15 consecutive epochs. 
We used the same implementation for all tasks and datasets while giving the same importance to each loss term in Equation~\ref{eq:overall_loss} without tuning, i.e., no specific task or dataset optimization is performed.

For both learning paradigms, model selection is performed on the training sets with the lowest loss value similar to~\cite{the_way_to_my_heart,demirel2023chaos}.
The reported results are mean and standard deviation values across three independent runs with different random seeds. 

\paragraph{Supervised}
We follow the same implementation as previous works~\cite{KDD_paper, demirel2023chaos} for supervised learning in time series. 
Specifically, we use two different network architectures: the first is a combination of convolutional and LSTM-based networks~\cite{KDD_paper, CorNET}, similar to the self-supervised implementation of the encoder architecture, and the second is a fully convolutional neural network that includes a final linear layer in the end instead of a projection block.
For the training of both architectures, we use categorical cross entropy as the loss function, which is optimized using Adam with a learning rate of $0.001$.
We train the models with a batch size of 64 for 120 epochs.
The model selection is performed on the validation sets with the lowest loss, where the validation set is created by randomly splitting 10\% of remaining data after excluding the test subjects.
Details of the architectures and parameter comparisons are provided in Appendix~\ref{appendix:arch_details}. 
The performance of various architectures with different sizes is discussed in Appendix~\ref{appendix:supervised_arch}.

\paragraph{Traditional}
We also compare our method with the rule-based periodicity detection methods in addition to the learning-based techniques.
In our evaluation, we consider the frequency domain approach, which detects underlying periodic patterns using Fourier transforms, and the time domain, where signals are correlated with themselves through autocorrelation function (ACF)~\cite{ICASSP_23}.
Additionally, we compare our method with an approach that combines the time and frequency domains using the periodogram and the ACF, integrating the Fisher method for periodicity detection~\cite{robustperiod}.
Since these methods are not learning-based, they fail to detect periodicities when evaluated on datasets containing periodic noise in the same band, limiting generalization across various conditions.
Since traditional methods lack randomness, we conducted the experiments only once and reported the results, i.e., no standard deviation is calculated for rule-based methods.

%%%%%%%%%%%%%%%%%%%%% ALL appendix %%%%%%%%%%%%%%%%%%%%%%%%%

% \begin{equation}\label{eq:MAE}
%     \text{MAE} = \frac{1}{W} \sum_{w=1}^{W} |\text{HR}_{est}(w) - \text{HR}_{ref}(w)| ,
% \end{equation}

% where W is the total number of segments, and $\text{HR}_{est}(w)$ and $\text{HR}_{ref}(w)$ denote the estimated and reference heart rate value in beats-per-minute on the $w^{th}$ segment, respectively. 
% This performance metric is widely used in PPG-based HR estimation~\cite{DeepPPG, sch_original}. 
% The $\text{HR}_{est}(w)$ is obtained using our model and the $\text{HR}_{ref}(w)$ values are directly used from datasets that share HR values corresponding to segments~\cite{DeepPPG, TROIKA}. 
% For the datasets that HR values are not available for the segmented PPG signals~\cite{WESAD}, we extracted HR information from ECG signals using Pan-Tompkins algorithm~\cite{Pan_Tompkins}.
\section{Results and Discussion}
\label{ref:results}
We present the main results of our proposed approach compared to state-of-the-art methods across the three time series tasks in eight datasets in Tables~\ref{tab:performance_ecg} to~\ref{tab:performance_ppg}.
Overall, the proposed set of regularizers has demonstrated substantial performance improvements, reaching up to 45--50\% in some tasks, compared to other supervised and self-supervised learning methods with rule-based techniques.
More importantly, our method outperforms other supervised learning-based techniques, which require label information, while eliminating the necessity for implementing specialized data augmentation methods to learn periodic representations from data.

Tables~\ref{tab:performance_resp} and~\ref{tab:performance_ppg} show that the rule-based methods fail to detect periodicity when applied to datasets containing multiple periodic components with noise. 
For example, autocorrelation, one of the most common methods for periodicity detection~\cite{autoformer, cyclo_detection}, performs poorly in various datasets that contain multiple periodicities with a high noise where the periodicity of interest is hidden.

Another interesting observation from the results is that the SimPer, designed for extracting representations related to periodic information from data similar to this work, performs worse when applied to noisy time series compared to other self-supervised techniques, despite sharing the same downstream task.
This might be attributed to the fact that SimPer designed specialized augmentations that alter the periodicity of the sample within and contrast through that for representation learning. 
However, when dealing with time series containing multiple periodicities under noisy conditions, the specific augmentation can lead to spectral overlapping, resulting in suboptimal performance.
Moreover, since there is no label information during self-supervised training, this method can also learn the features related to periodic noise, which is impossible to filter out as it lies on the same frequency band with the interested periodicity.

From the results, we can see that our proposed method significantly outperforms SiNC and VICReg, which impose a hinge loss on the variance over a batch of representations to enforce diverse outputs to prevent collapse, by a large margin (up to 73.9\% with a 10.1\% on average in tasks).

\begin{table}[t!]
\caption{Comparison of methods in \textit{Step counting}}
\begin{adjustbox}{width=1\columnwidth,center}
\label{tab:performance_step}
\renewcommand{\arraystretch}{0.7}
\begin{tabular}{@{}lllll@{}}
\toprule
\multirow{2}{*}{Method} & \multicolumn{2}{l}{Clemson (Regular)} & \multicolumn{2}{l}{Clemson (Semi-regular)} \\ 
\cmidrule(r{15pt}){2-3}  \cmidrule(r{5pt}){4-5}  \\ 
& MAPE$\downarrow$ & MAE$\downarrow$ &  MAPE$\downarrow$ & MAE$\downarrow$ \\
\midrule
\textit{Heuristic} & & & &  \\
Fourier & 12.16 & 7.60 & 44.94 & 13.23 \\
Autocorrelation & 39.96 & 24.45 & 47.64 & 13.76 \\
RobustPeriod & 14.87 & 24.20 & 47.39 & 13.91 \\
\midrule
\textit{Supervised} & & & &  \\
DCL & 5.99\small$\pm$0.34 & 3.45\small$\pm$0.32 & 19.59\small$\pm$1.54 & 8.98\small$\pm$0.87  \\
FCN & 6.15\small$\pm$0.60 & 3.53\small$\pm$0.33 & \textbf{17.22}\small$\pm$0.31 & \textbf{6.98}\small$\pm$0.23 \\
\midrule
\textit{Self-Supervised} & & & & \\
SimCLR & 6.22\small$\pm$0.17 & 5.52\small$\pm$0.14 & 24.41\small$\pm$1.80 & 14.22\small$\pm$1.52 \\
NNCLR & 6.26\small$\pm$0.81 & 5.58\small$\pm$0.72 & 25.12\small$\pm$0.74 & 15.38\small$\pm$0.58  \\
BYOL & 6.69\small$\pm$0.54 & 5.95\small$\pm$0.48 & 25.48\small$\pm$0.46 & 15.22\small$\pm$0.30  \\
TS-TCC & 6.01\small$\pm$0.33 & 3.80\small$\pm$0.38 & 20.08\small$\pm$0.33 & 10.95\small$\pm$0.33 \\
TS2Vec & 6.54\small$\pm$0.76 & 5.17\small$\pm$0.64 & 26.80\small$\pm$0.60 & 16.62\small$\pm$0.46 \\
VICReg & 10.93\small$\pm$3.96 & 9.81\small$\pm$3.52 & 26.90\small$\pm$1.02 & 15.92\small$\pm$0.84 \\
Barlow Twins & 6.87\small$\pm$0.16 & 5.46\small$\pm$0.13 & 19.97\small$\pm$0.70 & 11.14\small$\pm$0.46 \\
SimPer & 7.03\small$\pm$0.57 & 6.75\small$\pm$1.43 & 23.43\small$\pm$0.95 & 15.12\small$\pm$0.41 \\
\midrule
\textit{No Supervision} & & & & \\
SiNC & 22.92\small$\pm$6.61 & 13.83\small$\pm$4.02 & 37.12\small$\pm$0.01 & 12.31\small$\pm$0.28\\
Ours & \textbf{5.95}\small$\pm$0.21 & \textbf{3.42}\small$\pm$0.17 & 35.18\small$\pm$0.02 & 14.21\small$\pm$0.97  \\
\bottomrule
\end{tabular}
\end{adjustbox}
\end{table}
These empirical findings suggest that the assumption of having diverse samples in a batch is quite arguable, especially depending on the downstream task, where the pattern of interest can be severely hidden by noise such that the model can learn random features from the samples to have diversity in a batch of representations.
The empirical results align with our initial proposition and the motivation behind introducing a novel regularizer term, which prevents the model collapse by ensuring that the extracted representations retain traces from the samples from which they originated rather than relying on random variance constraints in the batch.
\begin{table}[t!]
\caption{Comparison of methods in \textit{Respiration}}
\begin{adjustbox}{width=1\columnwidth,center}
\label{tab:performance_resp}
\renewcommand{\arraystretch}{0.63}
\begin{threeparttable}
\begin{tabular}{@{}lllll@{}}
\toprule
\multirow{2}{*}{Method} & \multicolumn{2}{l}{CapnoBase (64-second)} & \multicolumn{2}{l}{CapnoBase (32-second)} \\ 
\cmidrule(r{7pt}){2-3}  \cmidrule(r{7pt}){4-5}  \\ 
& MAE$\downarrow$ & RMSE$\downarrow$ &  MAE$\downarrow$ & RMSE$\downarrow$ \\
\midrule
\textit{Heuristic} & & & &  \\
Fourier & 4.08 & 4.96 & 4.66 & 5.47 \\
Autocorrelation & 23.08 & 26.65 & 36.77 & 40.45 \\
RobustPeriod & 7.94 & 8.50 & 8.03 & 8.55 \\
\midrule
\textit{Supervised} & & & &  \\
DCL & 5.76\small$\pm$0.28 & 7.45\small$\pm$0.27 & 5.74\small$\pm$0.08 & 7.68\small$\pm$0.07  \\
FCN & 6.00\small$\pm$0.20 & 8.15\small$\pm$0.28 & 5.41\small$\pm$0.24 & 7.57\small$\pm$0.40 \\
\midrule
\textit{Self-Supervised} & & & & \\
SimCLR & 3.72\small$\pm$0.27 & 4.94\small$\pm$0.20 & 3.93\small$\pm$0.70 & 5.48\small$\pm$1.00 \\
NNCLR & 3.70\small$\pm$0.52 & 4.83\small$\pm$0.45 & 3.86\small$\pm$0.69 & 5.58\small$\pm$1.06  \\
BYOL & 4.09\small$\pm$0.40 & 5.45\small$\pm$0.42 & 4.40\small$\pm$1.19 & 6.05\small$\pm$1.58  \\
TS-TCC & 4.37\small$\pm$0.45 & 5.74\small$\pm$0.41 & 4.74\small$\pm$1.60 & 6.38\small$\pm$1.84 \\
TS2Vec & 7.00\small$\pm$3.30 & 8.44\small$\pm$4.62 & 5.57\small$\pm$2.76 & 7.93\small$\pm$3.83 \\
VICReg & 3.99\small$\pm$0.43 & 5.21\small$\pm$0.42 & 4.05\small$\pm$0.83 & 5.77\small$\pm$1.19 \\
Barlow Twins & 3.56\small$\pm$0.43 & 4.76\small$\pm$0.50 & 4.21\small$\pm$1.01 & 5.85\small$\pm$1.40 \\
SimPer & 3.89\small$\pm$0.64 & 5.03\small$\pm$1.26 & 5.01\small$\pm$0.90 & 7.96\small$\pm$1.75 \\
\midrule
\textit{No Supervision} & & & & \\
SiNC & 4.11\small$\pm$0.10 & 5.44\small$\pm$0.31 & 4.09\small$\pm$0.04 & 5.29\small$\pm$0.06 \\
Ours & \textbf{3.40}\small$\pm$0.20 & \textbf{4.41}\small$\pm$0.43 & \textbf{3.74}\small$\pm$0.03 & \textbf{4.77}\small$\pm$0.12  \\
\bottomrule
\end{tabular}
  \end{threeparttable}
\end{adjustbox}
\end{table}

\begin{table*}[t]
\caption{Performance Comparison of ours with other methods in \textit{PPG} datasets for HR estimation}
\begin{adjustbox}{width=2\columnwidth,center}
\label{tab:performance_ppg}
\renewcommand{\arraystretch}{0.8}
\begin{tabular}{@{}lllllllllll@{}}
\toprule
\multirow{2}{*}{Method} & \multicolumn{3}{l}{IEEE SPC12} & \multicolumn{3}{l}{IEEE SPC22} & \multicolumn{3}{l}{DaLiA$_{PPG}$} \\ 
\cmidrule(r{15pt}){2-4}  \cmidrule(r{15pt}){5-7}  \cmidrule(r{15pt}){8-10} \\ 
& MAE$\downarrow$ & RMSE$\downarrow$ & $\rho$$\uparrow$ & MAE$\downarrow$ & RMSE$\downarrow$ & $\rho$ $\uparrow$ & MAE$\downarrow$ & RMSE$\downarrow$ & $\rho$$\uparrow$ \\
\midrule
\textit{Heuristic} & & & & & & & & & \\
Fourier & 12.12 & 20.09 & 72.17 & 14.83 & 25.81 & 35.15 & 34.98 & 47.13 & 2.72\\
Autocorrelation & 56.65 & 64.58 & 24.89 & 41.16 & 49.81  & 3.945 & 30.59 & 39.09 & 9.18 & \\
RobustPeriod & 19.56 & 31.99 & 41.10 & 22.34 & 35.47 & 33.50 & 23.43 & 30.12 & 21.53 & \\
\midrule
\textit{Supervised} & & & & & & & & & \\
DCL & 19.90\small$\pm$1.10 & 26.34\small$\pm$1.10 & 25.53\small$\pm$4.2 & 22.43\small$\pm$0.62 & 27.17\small$\pm$0.60 & 11.95\small$\pm$5.1 & \textbf{5.97}\small$\pm$0.19& \textbf{11.83}\small$\pm$0.36 & \textbf{80.41}\small$\pm$0.9\\
CNN & 10.05\small$\pm$0.18 & 18.12\small$\pm$0.40 & 67.20\small$\pm$0.9 & 17.97\small$\pm$0.33 & 23.06\small$\pm$0.36 & 21.91\small$\pm$1.76 & 7.35\small$\pm$0.14 & 13.74\small$\pm$0.26 & 74.22\small$\pm$0.55 \\
\midrule
\textit{Self-Supervised} & & & & & & & & & \\
SimCLR & 12.42\small$\pm$0.05 & 20.96\small$\pm$0.30 & 60.41\small$\pm$0.52 & 21.08\small$\pm$1.79 & 27.94\small$\pm$3.17&13.20\small$\pm$9.97 & 12.01\small$\pm$0.14 & 19.46\small$\pm$0.14 &58.31\small$\pm$0.39 & \\
NNCLR & 13.14\small$\pm$0.49 & 18.86\small$\pm$0.49 & 69.82\small$\pm$0.06 & 20.79\small$\pm$0.90 & 26.39\small$\pm$1.45 & 14.73\small$\pm$6.34 & 12.94\small$\pm$0.31 & 20.02\small$\pm$0.49 & 51.12\small$\pm$2.54 \\
BYOL & 18.71\small$\pm$0.93 & 25.01\small$\pm$1.50 & 48.82\small$\pm$4.36 & 20.01\small$\pm$0.80 & 25.58\small$\pm$1.19 &15.25\small$\pm$1.45 & 11.67\small$\pm$0.32 & 17.57\small$\pm$0.23 &63.96\small$\pm$0.97 & \\
TS-TCC & 11.56\small$\pm$0.41 & 18.04\small$\pm$0.66 & 68.38\small$\pm$1.41 & 18.77\small$\pm$0.23 & 23.90\small$\pm$0.31 & 28.73\small$\pm$1.40 & 8.12\small$\pm$0.30 & 14.89\small$\pm$0.21 & 67.13\small$\pm$0.53 \\
TS2Vec & 9.75\small$\pm$0.08 & 17.82\small$\pm$0.43 & 75.43\small$\pm$0.33 & 25.77\small$\pm$0.17 & 31.90\small$\pm$0.16 & 06.40\small$\pm$2.31 & 10.83\small$\pm$0.13 & 17.89\small$\pm$0.19 & 60.10\small$\pm$0.62 \\
VICReg & 13.17\small$\pm$0.82 & 20.38\small$\pm$1.27 & 59.76\small$\pm$4.16 & 24.17\small$\pm$0.62 & 30.53\small$\pm$0.41 & 15.28\small$\pm$3.45 & 14.90\small$\pm$0.16 & 21.94\small$\pm$0.11 & 45.38\small$\pm$0.07
\\
Barlow Twins & 13.22\small$\pm$0.34 & 20.42\small$\pm$0.88 & 64.51\small$\pm$4.01 & 25.14\small$\pm$0.89 & 31.35\small$\pm$0.82 & 10.15\small$\pm$2.04 & 18.26\small$\pm$0.57 & 23.41\small$\pm$0.29 & 23.42\small$\pm$7.30

\\
SimPer & 15.10\small$\pm$0.20 & 21.20\small$\pm$0.26 & 52.75\small$\pm$0.84 & 26.30\small$\pm$0.35 & 30.90\small$\pm$0.63 & 05.30\small$\pm$3.38 & 16.35\small$\pm$0.36 & 23.00\small$\pm$0.63 & 38.11\small$\pm$1.38 \\
\midrule
\textit{No Supervision} & & & & & & & & & \\
SiNC & 19.34\small$\pm$5.38 & 28.44\small$\pm$5.41 & 38.35\small$\pm$7.60 & 21.93\small$\pm$5.46 & 25.52\small$\pm$5.18 & 24.61\small$\pm$6.20 & 14.20\small$\pm$1.30 & 25.66\small$\pm$1.39 & 27.91\small$\pm$4.65 & \\
Ours & \textbf{9.30}\small$\pm$0.10 & \textbf{16.50}\small$\pm$0.20 & \textbf{77.60}\small$\pm$0.43 & \textbf{10.27}\small$\pm$0.37 & \textbf{19.62}\small$\pm$0.71 & \textbf{44.10}\small$\pm$0.89 & 27.41\small$\pm$4.73 & 31.26\small$\pm$4.55 & 18.12\small$\pm$3.86 & \\
\bottomrule
\end{tabular}
\end{adjustbox}
\end{table*}
Another interesting outcome of the results is when our method is outperformed by supervised and self-supervised methods while having a minor performance gap with unsupervised techniques, which is the case for DaLiA$_{PPG}$ and Clemson semi-regular.
We conjecture that the performance difference between supervised and unsupervised methods in these datasets is due to the average level of noise in the samples, making the periodic pattern of interest completely hidden and hard to extract without supervision.
Similarly, in the case of semi-regular walking, the periodic patterns in the training decrease significantly compared to regular, leading to poor performance by unsupervised methods due to their inability to effectively learn desired patterns.
Since our approach learns the function $f_{\theta}$ by minimizing the expectation of the proposed loss, which is known as the empirical risk minimization (ERM), the model converges to a point where the risk is minimal in the training set.
When the process is unsupervised, the model might inadvertently learn the periodic noise or random features from the data, which leads to failures during the evaluation.
Contrarily, in self-supervised and supervised scenarios, models may learn dataset statistics with label information, such as mean values.
The model can then use this during evaluation, contributing to improved performance for imbalanced datasets~\cite{class_imbalance}.

These results suggest that the methods should be evaluated with several tasks in multiple datasets to investigate if they generalize and perform well under different characteristics and conditions, i.e., noise types and levels, narrow/wide bandwidth for the periodicity of interest.
Otherwise, the hyperparameters (specific augmentations with thresholds) can be optimized for a single task while performing poorly for the rest.
It is important to emphasize that our implementation (the model architecture and weights for loss terms) remained consistent across all experiments, ensuring a standardized evaluation.
We conduct detailed ablation experiments to investigate further the impact of components of our method on the performance across the datasets.
\setlength{\textfloatsep}{15pt plus 1.0pt minus 4.0pt}
\begin{figure}[h]
  \begin{subfigure}{.24\textwidth}
  \centering
    \includegraphics[width=\linewidth]{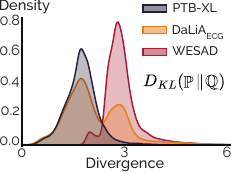}
    \caption{HR from ECG}
  \end{subfigure}%
  \begin{subfigure}{.24\textwidth}
  \centering
    \includegraphics[width=\linewidth]{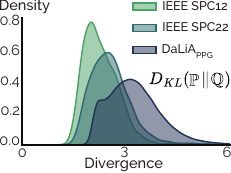}
    \caption{HR from PPG}
  \end{subfigure}
  \begin{subfigure}{.24\textwidth}
  \centering
    \includegraphics[width=\linewidth]{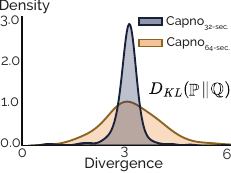}
    \caption{Respiratory}
  \end{subfigure}%
  \begin{subfigure}{.24\textwidth}
  \centering
    \includegraphics[width=\linewidth]{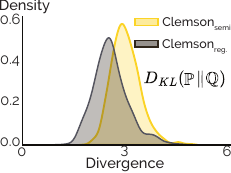}
    \caption{Step counting}
  \end{subfigure}
  \caption{The distribution of the spectral relative entropies between samples $\mathbb{P}$ and optimal waveforms $\mathbb{Q}$ which are obtained for each sample using the labels.
  In other words, $S_X(\omega) \sim \mathbb{P}$, and $S_{f^*_{\theta}}(\omega) \sim \mathbb{Q}$, where $f^*_{\theta}$ is the optimal model that extracts representations that only contain the desired periodicity.}
  \label{fig:average_KL_divergence}
\end{figure}
\subsection{Ablation Study}
\label{sec:ablation}
In this section, we present a comprehensive examination of our proposed method and the effect of its components on performance of the models for detecting the periodicity of interest.
Mainly, we investigate the effect of each proposed loss term on performance across datasets while establishing connections between dataset characteristics in relation to the average noise levels and learning paradigms.
Therefore, first, we extend our investigation for the case when the supervised algorithms significantly outperform our approach.
Figure~\ref{fig:average_KL_divergence} illustrates the distribution of relative entropies computed between the spectra of original samples and optimal waveforms using a continuous probability density curve~\cite{seaborn}, where the optimal samples only contain a pure sinusoidal at the frequency of the interest.

\begin{table*}[h]
\centering
\caption{Ablation on proposed regularizers in \textit{PPG} datasets for HR estimation}
\begin{adjustbox}{width=2\columnwidth,center}
\label{tab:performance_ppg_ablation}
\renewcommand{\arraystretch}{0.7}
\begin{tabular}{@{}lllllllllll@{}}
\toprule
\multirow{2}{*}{Method} & \multicolumn{3}{l}{IEEE SPC12} & \multicolumn{3}{l}{IEEE SPC22} & \multicolumn{3}{l}{DaLiA$_{PPG}$} \\ 
\cmidrule(r{15pt}){2-4}  \cmidrule(r{15pt}){5-7}  \cmidrule(r{15pt}){8-10} \\ 
& MAE$\downarrow$ & RMSE$\downarrow$ & $\rho$$\uparrow$ & MAE$\downarrow$ & RMSE$\downarrow$ & $\rho$ $\uparrow$ & MAE$\downarrow$ & RMSE$\downarrow$ & $\rho$$\uparrow$ \\
\midrule
Combined ($\mathcal{L}$) & \textbf{9.30}\small$\pm$0.10 & \textbf{16.50}\small$\pm$0.20 & \textbf{77.60}\small$\pm$0.43 & \textbf{10.27}\small$\pm$0.37 & \textbf{19.62}\small$\pm$0.71 & \textbf{44.10}\small$\pm$0.89 & 27.41\small$\pm$4.73 & 31.26\small$\pm$4.55 & 18.12\small$\pm$3.86 & \\
w/o Prop.~\ref{prop:spectral_entropy} & 12.69 (\textcolor{WildStrawberry}{-3.39})  & 20.92 (\textcolor{WildStrawberry}{-4.42}) & 71.01 (\textcolor{WildStrawberry}{-6.59}) & 14.51 (\textcolor{WildStrawberry}{-4.24}) & 25.30 (\textcolor{WildStrawberry}{-5.68}) & 38.03 (\textcolor{WildStrawberry}{-6.07}) & 18.71 (\textcolor{Green}{+4.7}) & 28.63 (\textcolor{Green}{+2.63}) & 31.13 (\textcolor{Green}{+10.0})  \\
w/o Prop.~\ref{prop:KL_divergence} & 39.14 (\textcolor{WildStrawberry}{-29.84}) & 46.01 (\textcolor{WildStrawberry}{-29.51})  & 15.95 (\textcolor{WildStrawberry}{-61.65}) & 39.71 (\textcolor{WildStrawberry}{-29.56}) & 42.66 (\textcolor{WildStrawberry}{-23.04}) & 0.771 (\textcolor{WildStrawberry}{-43.32}) & 35.26 (\textcolor{WildStrawberry}{-11.85}) & 40.13 (\textcolor{WildStrawberry}{-8.87}) & 10.03 (\textcolor{WildStrawberry}{-10.09})\\
Fourier & 12.12 (\textcolor{WildStrawberry}{-2.82}) & 20.09 (\textcolor{WildStrawberry}{-4.59}) & 72.17 (\textcolor{WildStrawberry}{-5.43}) & 14.83 (\textcolor{WildStrawberry}{-4.56}) & 25.81 (\textcolor{WildStrawberry}{-6.19}) & 35.15 (\textcolor{WildStrawberry}{-8.95}) & 34.98 (\textcolor{WildStrawberry}{-11.57}) & 47.13 (\textcolor{WildStrawberry}{-15.87}) & 2.72 (\textcolor{WildStrawberry}{-18.40})\\
\bottomrule
\end{tabular}
\end{adjustbox}
\end{table*}

\begin{table*}[hb!]
\centering
\caption{Ablation on proposed regularizers in \textit{ECG} datasets for HR detection}
\begin{adjustbox}{width=2\columnwidth,center}
\label{tab:performance_ecg_ablation}
\renewcommand{\arraystretch}{0.7}
\begin{tabular}{@{}lllllllllll@{}}
\toprule
\multirow{2}{*}{Method} & \multicolumn{3}{l}{PTB-XL} & \multicolumn{3}{l}{DaLiA$_{ECG}$} & \multicolumn{3}{l}{WESAD} \\ 
\cmidrule(r{15pt}){2-4}  \cmidrule(r{15pt}){5-7}  \cmidrule(r{15pt}){8-10} \\ 
& MAE$\downarrow$ & RMSE$\downarrow$ & $\rho$$\uparrow$ & MAE$\downarrow$ & RMSE$\downarrow$ & $\rho$$\uparrow$ & MAE$\downarrow$ & RMSE$\downarrow$ & $\rho$$\uparrow$ \\
\midrule
Combined ($\mathcal{L}$) & \textbf{3.75}\small$\pm$0.02 & \textbf{13.58}\small$\pm$0.03 & \textbf{79.30}\small$\pm$0.17 & \textbf{1.29}\small$\pm$0.001 & \textbf{2.08}\small$\pm$0.01 &\textbf{99.12}\small$\pm$0.01& \textbf{2.13}\small$\pm$0.01 & \textbf{3.88}\small$\pm$0.06 & \textbf{93.52}\small$\pm$0.61 & \\
w/o Prop.~\ref{prop:spectral_entropy} & 14.29 (\textcolor{WildStrawberry}{-10.54}) & 33.26 (\textcolor{WildStrawberry}{-19.6}) & 42.04 (\textcolor{WildStrawberry}{-37.26}) & 5.23 (\textcolor{WildStrawberry}{-3.94}) & 16.84 (\textcolor{WildStrawberry}{-14.76}) & 71.29 (\textcolor{WildStrawberry}{-27.83}) & 5.09 (\textcolor{WildStrawberry}{-2.96}) & 16.10 (\textcolor{WildStrawberry}{-12.2}) & 62.15 (\textcolor{WildStrawberry}{-31.3}) \\
w/o Prop.~\ref{prop:KL_divergence} & 7.74 (\textcolor{WildStrawberry}{-4.42}) & 21.82 (\textcolor{WildStrawberry}{-8.24}) & 46.73 (\textcolor{WildStrawberry}{-32.57}) & 2.93 (\textcolor{WildStrawberry}{-1.64}) & 5.89 (\textcolor{WildStrawberry}{-3.81}) &92.90 (\textcolor{WildStrawberry}{-6.22}) & 3.44 (\textcolor{WildStrawberry}{-1.31}) & 5.39 (\textcolor{WildStrawberry}{-1.51}) & 89.93 (\textcolor{WildStrawberry}{-4.59}) \\
Fourier & 10.51 (\textcolor{WildStrawberry}{-6.76}) & 28.65 (\textcolor{WildStrawberry}{-15.07}) & 49.12 (\textcolor{WildStrawberry}{-30.18}) & 4.06 (\textcolor{WildStrawberry}{-2.77}) & 15.46 (\textcolor{WildStrawberry}{-13.38}) & 73.73 (\textcolor{WildStrawberry}{-25.39}) & 4.88 (\textcolor{WildStrawberry}{-2.75}) & 18.57 (\textcolor{WildStrawberry}{-14.69}) & 56.69 (\textcolor{WildStrawberry}{-36.93})\\
\bottomrule
\end{tabular}
\end{adjustbox}
\end{table*}

Figure~\ref{fig:average_KL_divergence} gives an explanation for the performance variation in supervised and unsupervised learning across datasets by demonstrating and comparing the level to which the pattern of interest is hidden by noise.
Heuristic-based approaches typically require the SNR of a sample to exceed a certain threshold to detect the pattern, which can vary depending on the specific method, as these methods are not learning-based.
In contrast, our method is a learning-based approach designed to train a model that seeks to learn the patterns of interest using the training set and detect those patterns even when they are hidden by noise or other periodicities in unseen samples.
One significant drawback of relying on complete unsupervision for a learning-based method is that if the training set lacks sufficient diverse samples with a relatively high SNR for the model to learn the desired periodic patterns, the model may end up learning noise or random features from samples instead of the desired periodic pattern.
We, therefore, believe that the average SNR of a dataset is particularly important during the training of unsupervised and self-supervised learning methods (when there is no label information), as a higher average SNR indicates that relevant patterns, those related to downstream tasks, are distinguishable from noise.

Second, we analyze the contribution and impact of each proposed regularizer on model performance across various datasets under different levels and types of noise.
We, therefore, trained the neural network models with the same original implementation settings, i.e., the same architecture and training hyperparameters, using the different combinations of the loss components to analyze their contributions separately.
\begin{figure}[h]
  \begin{subfigure}{.24\textwidth}
  \centering
    \includegraphics[width=\linewidth]{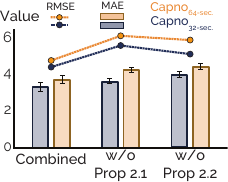}
    \caption{Respiratory}
    \label{fig:ablation_loss_resp}
  \end{subfigure}%
  \begin{subfigure}{.24\textwidth}
  \centering
    \includegraphics[width=\linewidth]{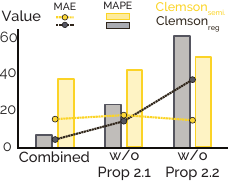}
    \caption{Step counting}
    \label{fig:ablation_loss_step}
  \end{subfigure}
    \caption{The ablation experiments on the effect of each loss term on the \textit{respiratory rate estimation}~Figure~\ref{fig:ablation_loss_resp} and \textit{step counting}~Figure~\ref{fig:ablation_loss_step} tasks.
    The x and y axes of both figures represent the loss combinations and corresponding error values for each task specific metric, respectively.
    The error decreases significantly when the model is trained using the combination of the proposed loss terms.
    The error bars for the second part are ignored as the deviation between random seeds is lower than 1--2\% of the overall error.}
  \label{fig:ablation_loss}
\end{figure}
Specifically, we first train the models without maximizing periodicity and while ensuring the extracted representations retain information to the original samples in the desired frequency band, i.e., $\mathcal{L}_1 = \mathcal{L}_{ds} + \mathcal{L}_{bw}$, without maximizing the periodicity (w/o Proposition~\ref{prop:spectral_entropy}, the second rows in Tables~\ref{tab:performance_ppg_ablation} and~\ref{tab:performance_ecg_ablation}).
Then, we train the models to maximize the periodicity without forcing the extracted representations to retain information from the original samples, i.e., $\mathcal{L}_2 = \mathcal{L}_{se} + \mathcal{L}_{bw}$, without preventing the collapse (w/o Proposition~\ref{prop:KL_divergence}, the third rows in Tables~\ref{tab:performance_ppg_ablation} and~\ref{tab:performance_ecg_ablation}). 
We also add the bandwidth loss, a common practice in the literature, to the loss variants for each case.
The experimental results for each dataset are presented in Tables~\ref{tab:performance_ppg_ablation} and~\ref{tab:performance_ecg_ablation} for heart rate detection task from PPG and ECG signals, respectively, and in Figure~\ref{fig:ablation_loss} for respiration and step counting tasks.

% \begin{table}[b]
% \caption{Comparison of methods in \textit{Step counting}}
% \begin{adjustbox}{width=1\columnwidth,center}
% \label{tab:performance_step_ablation}
% \renewcommand{\arraystretch}{0.5}
% \begin{tabular}{@{}lllll@{}}
% \toprule
% \multirow{2}{*}{Method} & \multicolumn{2}{l}{Clemson (Regular)} & \multicolumn{2}{l}{Clemson (Semi-regular)} \\ 
% \cmidrule(r{15pt}){2-3}  \cmidrule(r{5pt}){4-5}  \\ 
% & MAPE$\downarrow$ & MAE$\downarrow$ &  MAPE$\downarrow$ & MAE$\downarrow$ \\
% \midrule
% Combined ($\mathcal{L}$) & \textbf{5.95}\small$\pm$0.21 & \textbf{3.77}\small$\pm$0.17 & \textbf{35.18}\small$\pm$0.02 & \textbf{14.21}\small$\pm$0.97 
% \\
% w/o Prop.~\ref{prop:spectral_entropy} & 21.45 (\textcolor{WildStrawberry}{-15.5}) & 12.99 (\textcolor{WildStrawberry}{-9.22}) & 45.92 (\textcolor{WildStrawberry}{-10.7}) & 14.33 (\textcolor{WildStrawberry}{-0.1}) 
% \\
% w/o Prop.~\ref{prop:KL_divergence} & 54.99 (\textcolor{WildStrawberry}{-49.0}) & 33.47 (\textcolor{WildStrawberry}{-29.7}) & 38.73 (\textcolor{WildStrawberry}{-3.55}) & 16.21 (\textcolor{WildStrawberry}{-2.0}) \\
% \bottomrule
% \end{tabular}
% \end{adjustbox}
% \end{table}

When the models are trained without maximizing the periodicity in the extracted representations (w/o Proposition~\ref{prop:spectral_entropy}), the performance of the models is similar to periodicity detection using the Fourier transformation where the strongest harmonic is detected from the frequency spectrum of samples. 
This behavior is expected since the applied regularizers transform the model into a non-adaptive bandpass filter with cut-off frequencies set according to the task.
The combination of the proposed loss terms consistently performs well in all tasks, especially when the models have sufficient examples with a high SNR to learn the underlying periodic patterns of interest in the training set.
However, when the training data is noisy, i.e., the majority of the training set includes samples with low SNR, where the desired patterns are completely hidden, the Fourier transformation method with bandpass performs well as can be seen in Table~\ref{tab:performance_ppg_ablation}. 

The results obtained from the ablation studies also support the previous theoretical propositions and our motivation for presenting a novel regularizer for preventing the degenerate solution, i.e., the model collapse.
For example, a closer inspection of these results shows that when we train the models without relating representations with samples from which they originated, i.e., models that are trained only using $\mathcal{L}_1$ (w/o Proposition~\ref{prop:KL_divergence}), the models generally performed the worst among the three variants of the combinations, except for the ECG signals where the noise is less significant compared to other signal types, i.e., the periodic pattern of interest is more obvious.
This observation constitutes an empirical evidence for our proposed proposition, where a learner, which is only trained to maximize periodicity in uniformly sampled time series within a specific frequency band, can converge to a degenerate solution or learn the random waveforms in the signal.
Notably, this degenerated model even fails to identify obvious and trivial periodic patterns that can be detected by traditional rule based Fourier transformation or autocorrelation methods.

More ablations about the sensitivity of loss terms can be found in Appendix~\ref{appendix:loss_weights}.
Investigations regarding the effect of architectures in supervised learning methods are given in Appendix~\ref{appendix:supervised_arch}.
Comparative results regarding the performance of our approach with traditional methods that are specifically tailored for the applications are given in Appendix~\ref{appendix:more_traditional}.
Examples that visually demonstrate the learned representations from samples can be found in Appendix~\ref{appendix:figures_representations}.
Appendix~\ref{appendix:efficiency} provides a detailed evaluation and comparison of the computational efficiency of our proposed method.

\section{Related Work}
\label{sec:related_work}

\paragraph{Periodicity Detection}
Numerous studies have investigated periodicity detection for several time series applications~\cite{saul_periodic, robert_knight_periodic}.
Detecting the periodicity of physiological signals, such as heart and respiratory rate, plays a pivotal role in understanding individuals' health conditions~\cite{penzel_dynamics_2003}. 
Periodicity detection also finds application in workload forecasting~\cite{yu2024amortizedperiod}, anomaly detection~\cite{robustperiod}, and voice/speech analysis~\cite{saul_periodic}.
For instance, it helps identify features like pitch and intonation, which convey essential aspects of human communication, including recognizing and understanding speech~\cite{mit_acoustic}.
Although current methods are valid in addressing specific problems, a common characteristic is the development of specialized tools tailored to particular issues with optimized hyperparameters, i.e., several thresholds~\cite{robustperiod}, often lacking generalized solutions under noisy data~\cite{saul_periodic}.
Moreover, if there are two periodicities in the interested frequency bandwidth (i.e., noise and pattern of interest), these methods will fail to detect the desired one as they lack the adaptability to learn to distinguish between multiple periodic patterns~\cite{ICASSP_23}.
In contrast, our proposed method, which is adaptive and learns the desired pattern from the data in an unsupervised way, can detect periodicity across various conditions in three time series tasks without requiring labels for training or task-specific data augmentations with parameter optimizations.

\paragraph{Unsupervised Learning}
Self-supervised learning methods have gained significant attention as they enable the discovery of useful representations from data without labels by designing pretext tasks that change the unsupervised learning problem into a supervised one such as predicting the rotation of images~\cite{rotation_prediction}, or contexts~\cite{Context_prediction,Context_encoders}. 
Recently, SSL techniques, particularly in the vision domain, have been explored for periodicity detection from the sequence of frames~\cite{simper, Wang_Ahn_Kim_2022}.
Most of these methods~\cite{simper, the_way_to_my_heart} generate similar and dissimilar (positive/negative) pairs by applying tailored augmentations and learn periodic representations with slight modifications to the InfoNCE loss~\cite{InfoNCE}.
However, these approaches do not consider the challenges posed by noisy non-stationary time series data where signals may exhibit multiple periodicities in the interested bandwidth under significant noise~\cite{temko}.
Moreover, the usage of augmentations, in such cases, can be prone to issues like spectrum overlap, i.e., aliasing, resulting in the loss of information~\cite{oppenheim_2}.

Recent approaches have used regularizers to prevent collapse issue by penalizing the model when it produces similar embeddings within a batch~\cite{SiNC} instead of using similar/dissimilar samples as in contrastive learning.
For example, VICReg applies the hinge loss on the variance over a batch of predictions to enforce diverse representations/outputs~\cite{vicreg}.
However, having diverse samples in a batch is an overly optimistic claim, because during self-supervised training, where there is no label information available to the learners, this regularizer may punish the models for learning correct representations.
Even worse, the model might learn the noisy patterns to satisfy the arbitrarily set variance constraint in a batch.
Therefore, we take a different approach and introduce a new method to prevent the model collapse by ensuring the representations retain information from the original samples.
In other words, our approach hypothesizes that each representation should maintain a trace of the sample from which it is derived.
% Also, unlike existing methods that 
% BW based methods do not impose whitening constraints on the embedding, but they only require the embedding to be full-rank. This full-rank constraint is also sufficient to avoid dimensional collapse
% aim to generate hard samples—samples that are close to class boundaries—using adversarial approaches [53, 48, 72 ] or feature extrapolation [ 50, 71], our method seeks to connect semantically closer samples together using interpolation in a tailored way

\section {Conclusion}
\label{sec:Conclusion}
In this paper, we proposed a set of regularizers for detecting desired periodic patterns in time series sequences without requiring any label information or specialized data augmentation techniques.
Theoretically, we proved that an optimized learner trained with the proposed loss will detect the periodic pattern in the sequence while preventing degenerate solutions.
In contrast to previous methods that rely on optimistic assumptions on batch statistics to prevent the collapse of unsupervised models, we presented a novel approach that guarantees representations to preserve information from the original samples, thus relaxing assumptions on the batch.
Empirically, we showed that our method outperforms the prior works, achieving significant performance gains of up to 40--45\%, in three real-world tasks that involve diverse noisy conditions.
We believe that the methods introduced in this work have the potential to significantly improve learning solutions for a wide variety of time series tasks.

\section*{Impact Statement}
This paper aims to advance machine learning by introducing a new method for detecting periodic patterns in noisy time series data without relying on labeled data.
Since our method operates in an unsupervised manner, it can leverage vast amounts of unlabeled data commonly generated in everyday life.
This characteristic underscores its potential for real-world deployment, tapping into the rich resource of naturally occurring time series data for applications such as health monitoring, behavior analysis, and beyond.

\bibliography{sample}
\bibliographystyle{icml2024}

\newpage
\onecolumn
\appendix
% In the Appendix, we provide additional experiment details and spell out the proofs omitted in the text.

\section{Proof}
\label{appen:proof}
In this section, we present complete proofs of our theoretical study, starting with notations. 

\subsection{Notations and Preliminaries}
\label{appen:notations}
Fourier transform of a real-valued sample with a finite duration is obtained as in Equation~\ref{eq:appendix_fourier}.

\begin{equation}\label{eq:appendix_fourier}
    \mathcal{F}(\boldsymbol{\mathbf{x}}) = \sum_{n=-\infty}^{\infty} \boldsymbol{\mathrm{x}}_n e^{-j2\pi k n}
\end{equation}

The absolute frequency spectra of the samples are calculated as below.

\begin{equation}
    S_{X}(w) = \left| \sum_{n}^{} \mathbf{x}_n e^{-j\omega n}  \right|^2, \hspace{2mm} \text{where} \hspace{2mm} \omega = 2\pi f \hspace{2mm} \text{and} \hspace{2mm} f \in [0, f_s/2]
\end{equation}

We applied probability normalization to the obtained spectra for both original samples and the extracted representations in the desired frequency as in below before calculating the entropy and divergence.
\begin{equation}\label{appendix_eq:normalization}
\begin{gathered}
    S_X(\omega^*) = \frac{S_X(\omega^*)}{\sum_{\omega^*} S_{X}(\omega^*)}, \hspace{2mm} \text{where} \hspace{2mm} \omega^* \in  [0, 2 \pi f_s/2]
    \\[2pt]
    H(S_{X}) = \sum_{\omega} S_X(\omega^*) \log \frac{S_X(\omega^*)}{S_X(\omega^*)}
\end{gathered}
\end{equation}

The autocorrelation of a real-valued signal with time shift $\tau$ is calculated as in Equation~\ref{appendix:eq_autocorrelation}, where the denominator is the sample variance of the time series in that segment.

\begin{equation}\label{appendix:eq_autocorrelation}
    R_{XX}(\tau) = \frac{\sum_{n=1}^{N-\tau} (x_n - \mu_x)(x_{n+\tau} - \mu_x)}{\sum_{n=1}^{N} (x_n - \mu_x)^2}
\end{equation}

We also used the Wiener-Khintchine~\cite{Wiener, khintchine_korrelationstheorie_1934} theorem for calculating the spectral density of the sample $\mathbf{x}$ from the autocorrelation as in Equation~\ref{appendix:eq_wiener_khinchin}.

\begin{equation}\label{appendix:eq_wiener_khinchin}
    S(\omega) = \sum_{\tau}^{} R_{XX}(\tau) e^{-j\omega \tau}
\end{equation}

Similarly, as the time series samples are absolutely summable in our case, the autocorrelation can be obtained as in below.

\begin{equation}
    R_{XX}(\tau) = \frac{1}{2\pi} \int_{-\pi}^{\pi}S(\omega)e^{j \tau \omega}d(\omega)
\end{equation}

Since the spectral density is periodic in the frequency domain, the summation is performed over one period.

%%%%%%%%%%%%%%%%%%%%%%%%%%%%%%%%%%%%%%%%%%%%%%%%%%%%%%%%%%%%%%%%%%%%%%%%%%%%%%%%%%%%%%%%%%%%%%%%%%%%%%%%%%%%%%%%%%%%%%%%%%%
%%%%%%%%%%%%%%%%%%%%%%%%%%%%%%%%%%%%%%%%%%%%%%%%%%%%%%%%%%%%%%%%%%%%%%%%%%%%%%%%%%%%%%%%%%%%%%%%%%%%%%%%%%%%%%%%%%%%%%%%%%%
%%%%%%%%%%%%%%%%%%%%%%%%%%%%%%%%%%%%%%%%%%%%%%%%%%%%%%%%%%%%%%%%%%%%%%%%%%%%%%%%%%%%%%%%%%%%%%%%%%%%%%%%%%%%%%%%%%%%%%%%%%%
\subsection{Proof for Proposition~\ref{prop:spectral_entropy}}

\begin{proposition}[Maximizing periodicity]\label{appendix_prop:spectral_entropy}
Minimizing the spectral entropy of sequential data is nothing but maximizing its periodicity in the time domain.
\begin{equation}\label{appendix_eq:spectral_entropy_prop}
\begin{gathered}
\theta^* = \argmin_{\theta} \mathcal{L}_{se} = \argmax_{\theta} \mathcal{I}(\mathbf{y}; \mathbf{x}), 
\\
\text{where}  \hspace{2mm} \mathcal{L}_{se} = - \sum_{w}^{} S_{f_{\theta}}(w) \log S_{f_{\theta}}(w),
\\
S_{f_{\theta}}(w) = \frac{S_{f_{\theta}}(w)}{\sum_w S_{f_{\theta}}(w)}, \hspace{1mm} S_{f_{\theta}}(w) = \left| \sum_{n}^{} f_{\theta}(x)_n e^{-j\omega n}  \right|^2 
\end{gathered}
\end{equation}
\end{proposition}
\begin{proof}
The spectral entropy is maximum (periodicity is minimum) when the sequence generation process is a Gaussian with zero mean, $\sigma^2$ variance, i.e., $W(t) \sim \mathcal{N}(0, \sigma^2_w)$.
\begin{equation*}
    \mathbb{E}[W(t)] =0, \hspace{2mm} R_{WW}(\tau) = \mathbb{E}[W(t+\tau) W^*(t)]
\end{equation*}
For samples drawn from the Gaussian process, the correlation is non-zero only when $\tau = 0$ (as they are independent), and it is equal to the variance of the distribution. So;
\begin{gather}
R_{WW}(\tau) = 
    \begin{cases}
      \sigma^2_w, & \text{if } \tau = 0  \\    
      0, & \text{if } \tau \neq 0 
    \end{cases}
    \longrightarrow R_{WW}(\tau) = \sigma^2_w \delta(\tau)
\end{gather}
Using the Wiener-Khintchine theorem in Equation~\ref{appendix:eq_wiener_khinchin},
\begin{gather}
    S_{W}(\omega) = \sum_{\tau}^{} \sigma^2_w \delta(\tau) e^{-j \omega \tau}
    \\
    S_{W}(\omega) = 
    \begin{cases}
      \sigma^2_w, & \text{if } \omega \in (-\pi, \pi)  \\    
      0, & \text{otherwise}  
    \end{cases}
\end{gather}
In the last part of the above equation, the single period of the spectra in the frequency domain is considered for all samples.
When we apply the probability normalization to the obtained spectra in the desired frequency band,
\begin{gather}
    S_{W}(\omega) =\frac{1}{b-a}  \hspace{2mm} \text{where} \hspace{2mm} \omega \in  [a, b] \subset (-\pi, \pi)
\end{gather}
The obtained spectrum from the Gaussian process is nothing but the uniform distribution in the interested frequency band.
It is known that the Entropy is maximized if the distribution is uniform, therefore;
\begin{equation}\label{appendix_eq:proof_1_1_complete}
    H(S_{W}) \geq H(S_{X}), \forall \mathbf{x} \in X 
\end{equation}
The spectral entropy is minimum if the sequence generation process is periodic, i.e., $R_{PP}(\tau) = R_{PP}(\tau + T)$.
Since the $R_{PP}(\tau)$ is periodic with T, it can be expressed as a linear combination of harmonically related complex exponentials~\cite{Oppenheim}, which is also referred to as the Fourier series representation.
\begin{gather}
    R_{PP}(\tau) = \sum_{n=-\infty}^{\infty} \alpha_n e^{j \omega_0 n \tau}, \hspace{2mm} \omega_0 = \frac{2 \pi}{T}
    \\
    \alpha_n = \frac{1}{T} \int_{-T/2}^{T/2} R_{PP}(\tau) e^{-j \omega_0 n \tau} d\tau    
\end{gather}
Using the Wiener-Khintchine theorem in Equation~\ref{appendix:eq_wiener_khinchin}, we can write the spectrum density as follows;
\begin{gather}
    S_P(w) = \sum_{\tau}^{} R_{PP}(\tau) e^{-j2\pi \tau n}
    \\
    S_P(w) = \sum_{\tau}^{} \sum_{n}^{} \alpha_n e^{j \omega_0 n \tau} e^{-j \omega \tau}
    \\
    S_P(w) = \sum_{n}^{} \sum_{\tau}^{} \alpha_n e^{-j \tau (\omega-\omega_0 n)}
\end{gather}
Using the orthogonality of complex exponentials, we can write the above equation as follows,
\begin{gather}
S_P(w) = 
    \begin{cases}
      2\pi \sum_{n}^{} \alpha_n, & \text{for } \omega = \omega_0 n  \\    
      0, & \text{for } \omega \neq \omega_0 n 
    \end{cases}
    \longrightarrow S_P(w) = 2\pi\sum_{n-\infty}^{\infty} \alpha_n \delta(\omega-n\omega_0),
\end{gather}
where the $2\pi$ comes from the single period $(-\pi, \pi)$ summation of the autocorrelation.
When we apply the probability normalization to the obtained spectra in the desired frequency band while converting the discrete-time Fourier transform to the discrete Fourier transform\footnote{The Fourier transform is sampled at N equally spaced frequencies~\cite{Oppenheim}.}, we can obtain the following.
\begin{equation}\label{appendix_eq:S_P}
    S_P(w) = \delta(\omega-\omega_0) \hspace{2mm}  \text{where} \hspace{2mm} \omega \in  [a, b] \subset (0, \pi) 
\end{equation}
When we consider the positive frequencies, the normalized spectra will contain an impulse at the frequency of the periodicity, which is the case for minimum entropy.
\begin{gather}
    H(S_{P}) \leq H(S_{X}), \forall \mathbf{x} \in X  
\end{gather}
Combining Equation~\ref{appendix_eq:proof_1_1_complete} and the above, 
\begin{equation}
    H(S_{P}) \leq H(S_{X}) \leq H(S_{W})
\end{equation}
Therefore, $H(S_{P}) \leq \mathcal{L}_{se} \leq H(S_{W})$.
\end{proof}
The last inequality concludes the proof by showing that minimizing the spectral entropy of uniformly sampled time series data is equivalent to maximizing its periodicity.

%%%%%%%%%%%%%%%%%%%%%%%%%%%%%%%%%%%%%%%%%%%%%%%%%%%%%%%%%%%%%
%%%%%%%%%%%%%%%%%%%%%%%%%%%%%%%%%%%%%%%%%%%%%%%%%%%%%%%%%%%%%
\subsection{Proof for Proposition~\ref{prop:KL_divergence}}

\begin{proposition}[Degenerate solution]\label{appendix_prop:KL_divergence}
The relative entropy between the spectral distributions of samples and extracted representations is upper-bounded by a degenerate solution.
\begin{equation}\label{appendix_eq:KL_divergence}
\begin{gathered}
\infdiv{\mathbb{P}}{\mathbb{Q}} > \mathcal{L}_{ds} \geq 0, \hspace{2mm} \text{where}  \hspace{2mm} \text{Var}(\mathbb{Q}) = 0,
\\
\text{and}  \hspace{2mm} \mathcal{L}_{ds} = \sum_{\omega} S_X(\omega) \log \frac{S_X(\omega)}{S_{f_{\theta}}(\omega)},
\\
S_X(\omega) = \frac{S_X(\omega)}{\sum_{\omega} S_X(\omega)}, \hspace{1mm} S_X(\omega) = \left| \sum_{n}^{} x_n e^{-j \omega n}  \right|^2 
\end{gathered}
\end{equation}
\end{proposition}
\begin{proof}
From Equation~\ref{appendix_eq:S_P}, we can say that a learner, which is trained to maximize the periodicity, will output representations ($f_{\theta}(\mathbf{x})$) that have spectra as in below.
\begin{gather}
    S_{f_{\theta}}(\omega) = \delta(\omega-\omega_0)
\end{gather}
Nevertheless, this leads to an informational collapse, resulting in a degenerate solution where representations become identical, despite the maximum periodicity present.
In other words, if the distribution of representations is defined as $\mathbb{Q}$, the variance will be zero, i.e., $\text{Var}(\mathbb{Q}) = 0$.
Therefore,
% Therefore, we can say that there exists a spectrum $S_X(\omega)$ with a frequency $\hat{\omega}$ such that the spectrum has a non-zero value while the representation spectrum is zero.
\begin{gather}
    \exists \omega \in (-\infty, \infty), \quad S_{f_{\theta}}(\omega) = 0 \wedge S_X(\omega) \neq 0,
\end{gather}
And,
\begin{gather}
    D_{KL}(S_X(\omega) \hspace{1mm} \lVert \hspace{1mm} S_{f_{\theta}}(\omega)) = \sum_{\omega} S_X(\omega) \log \frac{S_X(\omega)}{S_{f_{\theta}}(\omega)} \rightarrow \infty
\end{gather}
Thus,
\begin{gather}
   D_{KL}(\mathbb{P} \hspace{1mm} \lVert \hspace{1mm} \mathbb{Q}) \geq  \mathcal{L}_{se} \geq 0
\end{gather}
% There exists a sample with a spectrum $S_X(\omega)$ for which a collapsed learner, trained to maximize periodicity (minimize spectral entropy), results in an infinite divergence of spectral distributions between samples and embeddings.
% \begin{gather}
%     \argmin_{\theta} \mathcal{L}_{se} \implies \lim_{\mathcal{L}_{se}\to 0} S_{f_{\theta}}(\omega) \to 0, \hspace{1mm} \exists \omega \in \mathcal{A}_{\omega}
%     \\
%     \exists \omega \in (-\infty, \infty), \quad S_{f_{\theta}}(\omega) = 0 \wedge S_X(\omega) \neq 0,
% \end{gather}
% yields $D_{KL}(\mathbb{P} \hspace{1mm} \lVert \hspace{1mm} \mathbb{Q})\hspace{-0.5mm}\rightarrow \hspace{-0.5mm} \infty$, \hspace{1mm} $S_X(\omega) \sim \mathbb{P}$, $S_{f_{\theta}}(\omega) \sim \mathbb{Q}$ 
\end{proof}
The proof suggests that degenerate solutions can be prevented if the learner seeks to minimize the relative entropy between the spectra of samples and extracted representations.
% -------------------------------------------------- %
% -------------------------------------------------- %
% -------------------------------------------------- %
% -------------------------------------------------- %
% -------------------------------------------------- %
% -------------------------------------------------- %
% -------------------------------------------------- %
% -------------------------------------------------- %
% -------------------------------------------------- %
% -------------------------------------------------- %
% -------------------------------------------------- %
% -------------------------------------------------- %
% -------------------------------------------------- %
% -------------------------------------------------- %
\section{Datasets}
\label{appendix:datasets}
In this section, we give details about the datasets that are used during our experiments.

\subsection{Heart rate}

\subsubsection{\textit{ECG} based}

\paragraph{PTB-XL}
The PTB-XL ECG is a large dataset of 21799 clinical 12-lead ECGs, where lead-II was used during experiments, from 18869 patients of 10-second length segments.
The dataset itself provides recommended splits into training and test sets.
We, therefore, follow the exact recommendation.
ten percent of the training training set is used as fine-tuning for self-supervised learning techniques.

\paragraph{DaLia$_{ECG}$}
PPG dataset for motion compensation and heart rate estimation in Daily Life Activities (DaLia) was recorded from 15 subjects (8 females, 7 males, mean age of $30.6$), where each recording was approximately two hours long. 
The signals were recorded while subjects went through different daily life activities, for instance sitting, walking, driving, cycling, working, and so on. 
ECG signals were recorded at a sampling rate of 700\,Hz. 
We follow leave-one-out-cross-validation for each subject.
For the self-supervised setting, we used the first five subjects for fine-tuning.

\paragraph{WESAD}
The multimodal WESAD data set includes physiological and mobility data from wrist-worn (Empatica E4) and chest-worn (RespiBAN) devices. 
Data were acquired from 15 subjects and contain multiple features, pulse rate, ECG, and body temperature, as extracted from a wrist-worn device, and blood volume pulse (BVP) and respiration extracted by the chest-worn devices~\cite{WESAD}.
We evaluate the dataset using leave-one-subject-out.

ECG datasets are standardized as follows.
Initially, a fourth-order Butterworth bandpass filter with a frequency range of 0.7--40\,Hz is applied to the signals. 
In the case of DaLia and WESAD, we use an 8-second sliding window with 2-second shifts for segmentation.
Differently, for PTB-XL, the ECG segments are pre-provided.
After segmentation, we calculated the square of the first differentiation before feeding samples to the models, which helps to emphasize beats~\cite{Pan_Tompkins}.
Lastly, the signals are normalized to zero to one range.

\subsubsection{\textit{PPG} based}

\paragraph{IEEE SPC}  This competition provided a training dataset of 12 subjects (SPC12) and a test dataset of 10 subjects~\cite{DeepPPG}. 
The IEEE SPC dataset overall has 22 recordings of 22 subjects, ages ranging from 18 to 58 performing three different activities~\cite{Binary_CorNet}. 
Each recording has sampled data from three accelerometer signals and two PPG signals along with a sampling frequency of 125\,Hz. 
All these recordings were recorded from the wearable device placed on the wrist of each individual. 
All recordings were captured with a 2-channel pulse oximeter with green LEDs, a tri-axial accelerometer, and a chest ECG for the ground-truth HR estimation. 
During our experiments, we used PPG channels. 
We use leave-one-out-cross-validation for the SPC12 and SPC22 as source domains similar to the previous setup while the last six subjects of SPC22 are used for source domains to prevent overlapping subjects with SPC12, similar to~\cite{demirel2023chaos}.

\paragraph{DaLia$_{PPG}$} 
PPG signals from the DaLia dataset were recorded at a sampling rate of 64\,Hz. 
The first five subjects are used as target domains with leave-one-out-cross-validation, following the~\cite{demirel2023chaos}.

All PPG datasets are standardized as follows.
Initially, a fourth-order Butterworth bandpass filter with a frequency range of 0.5--4\,Hz is applied to PPG signals. 
Subsequently, a sliding window of 8 seconds with 2-second shifts is employed for segmentation, followed by z-score normalization of each segment. 
Lastly, the signal is resampled to a frequency of 25\,Hz for each segment.

\subsection{Respiratory rate}

\paragraph{CapnoBase}
The CapnoBase dataset includes recordings of ECG and PPG along with capnometry signals from 42 subjects (13 adults, 29 children, and neonates).
The dataset also includes the inhaled and exhaled carbon dioxide (CO2) signal labeled by the research assistance~\cite{capno} and used as the reference breathing rate.
The dataset is split into 10 folds where each fold contains four subjects.
The last five fold is used as the test set for all baselines.
For the self-supervised learning, we used seven fold for pre-training and ten percent of the pre-training for fine-tuning.

\paragraph{BIDMC}
The BIDMC dataset consists of ECG, pulse oximetry (PPG), and impedance pneumography respiratory signals.
The data was acquired by randomly selecting critically ill patients undergoing hospital care at the Beth Israel Deaconess Medical Centre (Boston, MA, USA). 
Two trained annotators manually annotated individual breaths in each recording using the respiratory impedance signal. 
The data set contains 8-minute recordings of ECG, PPG, and impedance pneumography signals from 52 adult patients aged from 19 to older than 90, including 32 females~\cite{bidmc}.
For each set of annotations, the RR value was determined based on the average time between consecutive breaths within a given window; only those windows of data for which the agreement between both estimates was within 2 breaths per minute were retained.
We used 10-fold cross-validation for the evaluation where each fold contains 5 subjects.
For the self-supervised learning, we used seven fold for pre-training and ten percent of the pre-training for fine-tuning.

\subsection{Step counting}

\paragraph{Clemson}
The Clemson dataset has 30 participants, including 15 males and 15 females.
Each participant wore three Shimmer3 sensors.
We use the data from the device which is positioned on the non-dominant wrist.
The inertial measurement data is recorded at 15 Hz. 
The accelerometers were set to record from -2 to 2 gravities.
We computed the overall magnitude of the accelerometer and used it as input to the models.
We used the regular gait experiment, where the participant was instructed to walk two laps around a designated path at their normal walking pace. 
In the semi-regular gait experiment, participants were instructed to perform a scavenger hunt, locating four objects in four different rooms throughout a building~\cite{clemson}. 
We used 10-fold cross-validation for the evaluation where each fold contains 3 subjects.
For the self-supervised learning, we used seven fold for pre-training and two fold for fine-tuning.

When we split the datasets for training and testing folds, we ensure that each person's recordings appear in only one set.
We followed this procedure across all datasets to ensure that the trained model had not seen the testing data during training.

\subsection{Metrics}
We used the common evaluation metrics in the literature for each task.
Specifically, we used mean absolute error (MAE), root mean square error (RMSE), and Pearson correlation coefficient for \textit{heart rate prediction}~\cite{SiNC}, and we used the mean absolute percentage error (MAPE) for step counting~\cite{step_counting, FEMIANO2022206}.

Here, we explain how to calculate each metric for different time series tasks.
The MAE, RMSE, and correlation coefficient ($\rho$) are calculated using the following equations:

\begin{equation}\label{eq:MAE}
\text{MAE} = \frac{1}{n} \sum_{i=1}^{n} |\hat{y}_i - y_i|
\end{equation}

\begin{equation}\label{eq:RMSE}
\text{RMSE} =  \sqrt{\frac{\sum_{n=1}^{n} (\hat{y}_i - y_i)^2}{K}}
\end{equation}

\begin{equation}\label{eq:corr}
\rho = \frac{\sum_{i=1}^n (\hat{y}_i - \mu_{\hat{y}})(y_i - \mu_y)}{\sqrt{\sum_{i=1}^n (\hat{y}_i - \mu_{\hat{y}})^2}\sqrt{\sum_{i=1}^n (y_i - \mu_y)^2}}
\end{equation}

\begin{equation}\label{eq:MAPE}
    \text{MAPE} = \frac{100}{n} \sum_{i=1}^{n} \left|\frac{y_i - \hat{y}_i}{y_i}\right|\%
\end{equation}

where n represents the total number of segments. 
The variables $\hat{y}_i$ and $y_i$ denote the output of the model and ground truth values in beats (respiration)-per-minute or number of steps for the $n^{th}$ segment, respectively. 
We reported the percentage for the Pearson correlation coefficient to save precision in the tables.

\clearpage
% -------------------------------------------------- %
% -------------------------------------------------- %
% -------------------------------------------------- %
% -------------------------------------------------- %
% -------------------------------------------------- %
\section{Baselines}
\label{appendix:baselines}

\subsection{Traditional-Heuristic based methods}

\paragraph{Fourier transform}
We calculated the Fourier transform of signals and found the maximum amplitude sinusoidal from it to detect the periodicity.
The length of transformation is set to 2048 for ECG signals and 512 for the rest.
The frequency range of interest is defined according to common physiological limits for each task: [5, 40] \textit{rpm} for respiration rate, [40, 140] for step counts per minute, and [30, 210] \textit{bpm} for heart rate.
We increased the length of the Fourier transform for ECG signals as their sampling frequency is approximately four times higher than the other signals. 
During the loss calculation of our proposed method, we use the same hyperparameters, frequency range of interests with length, of Fourier transformation.

\paragraph{Autocorrelation}
Similar to the Fourier-based method, we calculate the autocorrelation, as in Equation~\ref{eq:autocorrelation}, of the signals and find the maximum value of $y_k$ in the range of frequency of interest.

\begin{equation}\label{eq:autocorrelation}
    y_k = \frac{\sum_{t=k+1}^{n}(x_t - \mu_x)(x_{t-k} - \mu_x)}{\sum_{t=1}^{n}(x_t - \mu_x)^2}
\end{equation}

\paragraph{RobustPeriod}
RobustPeriod adopts maximal overlap discrete wavelet transform (MODWT) to decompose the input time series into multiple time series at different levels, ranks by wavelet variance, and then performs Huber periodogram with autocorrelation.
We used Daubechies 10 wavelet with 4 levels, the lambda in the Hodrick–Prescott (HP) filter is set to 1e6, the Huber function hyperparameter is set to 2, and the M-Periodogram is set to 1.345.
Although we searched for the optimum hyperparameters, the search space was limited due to the significant time consumption arising from the wavelet computation and optimization for each sample, given the numerous hyperparameters involved in the technique.
Since this method includes autocorrelation to detect periodicity at each scale after the wavelet decomposition, the performance decreases severely when the desired period of interest is hidden by other periodic patterns in the same frequency band.

\subsection{Supervised}
\paragraph{DCL}
We used a similar implementation of DeepConvLSTM architecture which is a 4-layer convolutional neural network with a kernel size of five, and ReLU activation after each layer, followed by a Dropout~\cite{Dropout} and a two-layer LSTM with a hidden size being 128.
We chose this architecture for the supervised and self-supervised learning paradigms as it was widely used before in the literature~\cite{KDD_paper, demirel2023chaos}, especially for detecting the periodicities in the supervised paradigm from signals~\cite{CorNET}.

\paragraph{CNN}
We also implemented a fully convolutional neural network with a 3-layer followed by ReLU activation and MaxPooling after each convolutional layer.
Dropout with 0.5 is applied after the first convolutional layer. 
We set the kernel and padding size to 8 and 4, respectively for each convolutional layer. 
The number of kernels for each convolutional layer is set to 32 for the first one and 64 for the rest.

We also performed a grid search for the hyperparameters of the architectures where we mainly investigated the number of convolutional layers with the kernel size.
However, we did not observe a performance improvement with the increased number of parameters for the architectures as shown with the additional results in Appendix~\ref{appendix:supervised_arch}.

\subsection{Self-Supervised}
\label{appendix:baselines_SSL}

\paragraph{NNCLR}
We follow a similar setup to SimCLR by applying two separate data augmentations, then we use nearest neighbors in the learned representation space as the positive in contrastive losses~\cite{NNCLR}.
The maximum size of the support set equals 1024.

\paragraph{BYOL}
For the BYOL implementation, the exponential moving average parameter is set to 0.996 where the projector size is set to 128.
We set the learning rate to 0.03 similar to other SSL techniques.
Following the original implementation, we use a weight decay parameter of $1.5e-6$.

\paragraph{TS-TCC}
We follow the same architecture implementation with the losses, i.e., contextual and temporal contrasting.
TS-TCC proposed applying two separate augmentations, one augmentation is weak (jitter-and-scale) and the other is strong (permutation-and-jitter). 
The authors also change the strength of the permutation window from dataset to dataset. 
In our experiments, we first used the original augmentations for each time series task, however, we observed performance decreases depending on the signal type.
We, therefore, applied the specific augmentations for each time series, where we observed a general performance improvement in other SSL techniques as well.

\paragraph{TS2Vec} 
TS2Vec~\cite{TS2VEC} is a SSL method specifically designed for time series based on contrastive (instance and temporal wise) learning in a hierarchical way over augmented context views where the context is generated by applying timestamp masking and random cropping on the input time series.
Following the original framework, we use a dilated CNN architecture with a depth of 10 and hidden size of 64, which has a similar number of parameters with the architectures used by other SSL methods. 
The batch size is set to 256, and the number of epochs to 120, consistent with other SSL techniques.

\paragraph{VICReg}
We follow the original implementation and set the coefficients for each loss term to 25 ($\lambda$), 25 ($\mu$), and 1 ($\nu$), corresponding to the invariance, variance, and covariance terms, respectively.
Although we conducted a search for these loss term values, no performance enhancements were detected across the tasks.

\begin{equation}
    \ell = \lambda \left[ s(z,z^{\prime})\right] + \mu \left[ v(z) + v(z^\prime) \right] + \nu \left[  c(z) + c(z^{\prime}) \right],
\end{equation}

where $s$ is the mean-squared Euclidean distance, $v$ is a hinge function on the standard deviation of the embeddings along the batch dimension, $c$ is the covariance regularization term as the sum of the squared off-diagonal coefficients

\paragraph{Barlow Twins}
Barlow Twins~\cite{barlow_twins, barlow_real} presents an objective function that naturally avoids collapse for SSL by measuring the cross-correlation matrix between the outputs of two identical networks fed with augmented versions of a sample, and making it as close to the identity matrix as possible. 
This causes the embedding vectors of augmented versions of a sample to be similar, while minimizing the redundancy between the components of these vectors.
Following the original implementation, we applied batch normalization to the extracted embeddings and set the hyperparameter $\lambda$ coefficient (in Equation~\ref{eq:barlow_twins}) to 0.005.

\begin{equation}\label{eq:barlow_twins}
\mathcal{L} = \sum_{i} (1 - C_{ii})^2 + \lambda \sum_{i} \sum_{j \neq i} C_{ij}^2,
\end{equation}

where \( C \) is the cross-correlation matrix computed between the two sets of normalized embeddings.

\paragraph{SimPer}
SimPer is a simple contrastive self-supervised technique for learning periodic information in data.
To exploit the periodic inductive bias, SimPer introduces customized augmentations and feature similarity measures. 
We follow the same augmentation in the original paper, and we downsample or interpolate the signals, which the authors call speed change.
The range and the number of periodicity variant frequency augmentation are searched over [0.5, 3] and [3, 20], respectively, as in the original paper.
We use the MXCorr for the similarity metric.
Although the objective of the SimPer is the same as our work, the performance is extremely poor in all tasks and datasets compared to our proposed method even though the SimPer requires labels for training.

\subsection{No supervision}

\paragraph{SiNC}
The three loss terms (bandwidth, sparsity, variance) with equal weights are implemented for signal estimation via the non-contrastive unsupervised learning (SiNC) technique. 
However, we adapted these limits individually for each task, following the specific frequency boundaries outlined in the Fourier transform.

\clearpage
% -------------------------------------------------- %
% -------------------------------------------------- %
% -------------------------------------------------- %
% -------------------------------------------------- %
% -------------------------------------------------- %
\section{Implementation Details}
\label{appendix:Implementation_Details}
Here, we have provided the details of the searched architectures, hyperparameters, and data augmentations to push the performance of the supervised and self-supervised techniques.
To begin, we considered the model topologies that were used in previous works~\cite{demirel2023chaos, KDD_paper, CorNET} and expanded upon them.
We also evaluate the 1D Res-Net~\cite{resnet1d} implementation; however, we observe no significant performance improvement across datasets. 
This aligns with findings from previous works indicating that complex models may perform worse for time series data~\cite{he2023domain}.     

\subsection{Architectures}
\label{appendix:arch_details}
Here, we present the details of architectures that are investigated for the performance of learning-based algorithms.
Mainly, we consider the baselines in the literature and increase the number of parameters approximately four times to observe if the performance increases.
Batch normalization~\cite{batch} is applied after each convolutional block, except the first version of DCL, similar to~\cite{KDD_paper}.
ReLU activation is employed following batch normalization, in line with~\cite{ResNet}, with the exception of the final activation function in the U-Net, which is set to hyperbolic tangent.
The number of parameters for each model is calculated for an input with 200 dimensions, i.e., a segment of the PPG signal.

\begin{table}[h]
    \setlength{\tabcolsep}{3pt} % Adjust this value to decrease/increase column spacing
    \caption{The model topologies for the combination of convolutional with LSTM-based networks. Table~\ref{appendix_tab_lstm_v1}) is the baseline architecture from previous works.
    Table~\ref{appendix_tab_lstm_v2}) is a similar model with $\approx$ 2x parameters.}
    \centering
    \begin{subtable}[t]{0.48\textwidth}
        \centering
        \caption{DeepConvLSTM Architecture}\label{appendix_tab_lstm_v1}
        \begin{tabular}{lll}
            \toprule
            Layer & \multirow{2}{*}{\begin{tabular}[l]{@{}l@{}}Kernel\\Size\end{tabular}} & \multirow{2}{*}{\begin{tabular}[l]{@{}l@{}}Output\\Size\end{tabular}} \\
            & & \\
            \midrule
            Input (1 channel) & - & (T, 1) \\
            Conv (64 kernels) & (5, 1) & (T-4, 64) \\
            Conv (64 kernels) & (5, 1) & (T-8, 64) \\
            Conv (64 kernels) & (5, 1) & (T-12, 64) \\
            Conv (64 kernels) & (5, 1) & (T-16, 64) \\
            Permute + Reshape & - & (64, T-16) \\
            Dropout (p=0.5) & - & (64, T-16) \\
            LSTM (2 layers, 128 units) & - & (128, T-16) \\
            Output (if backbone) & - & (128,) \\
            Linear (if not backbone) & - & (n\_classes,) \\
            \midrule
            \# Parameters & & $\approx$316k \\
            \bottomrule
        \end{tabular}
    \end{subtable}\hfill
    \begin{subtable}[t]{0.48\textwidth}
        \centering
        \caption{DeepConvLSTM\_{v2} Architecture}\label{appendix_tab_lstm_v2}
        \begin{tabular}{lll}
            \toprule
            Layer & \multirow{2}{*}{\begin{tabular}[l]{@{}l@{}}Kernel\\Size\end{tabular}} & \multirow{2}{*}{\begin{tabular}[l]{@{}l@{}}Output\\Size\end{tabular}} \\
            & & \\
            \midrule
            Input (1 channel) & - & (T, 1) \\
            Conv (64 kernels) & (7, 1) & (T-2, 64) \\
            Conv (128 kernels) & (5, 1) & (T-4, 128) \\
            Conv (128 kernels) & (5, 1) & (T-8, 128) \\
            Conv (256 kernels) & (5, 1) & (T-10, 256) \\
            Permute + Reshape & - & (256, T-10) \\
            Dropout (p=0.5) & - & (256, T-10) \\
            LSTM (2 layers, 128 units) & - & (128, T-10) \\
            Output (if backbone) & - & (128,) \\
            Linear (if not backbone) & - & (n\_classes,) \\
            \midrule
            \# Parameters & & $\approx$494k \\
            \bottomrule
        \end{tabular}
    \end{subtable}
\end{table}

\begin{table}[h]
    \setlength{\tabcolsep}{3pt} % Adjust this value to decrease/increase column spacing
    \caption{The model topologies for the fully convolutional neural networks. Table~\ref{appendix_tab_fcn_v1}) is the baseline architecture from previous works.
    Table~\ref{appendix_tab_fcn_v2}) is a similar model with $\approx$ 1.5x parameters.}
    \centering
    \begin{subtable}[t]{0.48\textwidth}
        \centering
        \caption{FCN Architecture}\label{appendix_tab_fcn_v1}
        \begin{tabular}{lll}
            \toprule
            Layer & \multirow{2}{*}{\begin{tabular}[l]{@{}l@{}}Kernel\\Size\end{tabular}} & \multirow{2}{*}{\begin{tabular}[l]{@{}l@{}}Output\\Size\end{tabular}} \\
            & & \\
            \midrule
            Input (1 channel) & - & (T, 1) \\
            Conv (32 kernels) & (8, 1) & (T, 32) \\
            MaxPool & (2, 1) & (T/2, 32) \\
            Dropout (p=0.2) & - & (T/2, 32) \\
            Conv (64 kernels) & (8, 1) & (T/2, 64) \\
            MaxPool & (2, 1) & (T/4, 64) \\
            Conv (128 kernels) & (8, 1) & (T/4, 128) \\
            MaxPool & (2, 1) & (T/8, 128) \\
            Output (if backbone) & - & (128,) \\
            Linear (if not backbone) & - & (n\_classes,) \\  
            \midrule
            \# Parameters & & $\approx$700k \\
            \bottomrule
        \end{tabular}
    \end{subtable}
    \begin{subtable}[t]{0.48\textwidth}
        \centering
        \caption{FCN\_v2 Architecture}\label{appendix_tab_fcn_v2}
        \begin{tabular}{lll}
            \toprule
            Layer & \multirow{2}{*}{\begin{tabular}[l]{@{}l@{}}Kernel\\Size\end{tabular}} & \multirow{2}{*}{\begin{tabular}[l]{@{}l@{}}Output\\Size\end{tabular}}  \\
            & & \\
            \midrule
            Input (1 channel) & - & (T, 1) \\
            Conv (32 kernels) & (8, 1) & (T, 32) \\
            Conv (32 kernels) & (8, 1) & (T, 32) \\
            MaxPool & (2,1) & (T/2, 32) \\
            Dropout (p=0.2) & - & (T/2, 32) \\
            Conv (64 kernels) & (6, 1) & (T/2, 64) \\
            Conv (64 kernels) & (6, 1) & (T/2, 64) \\
            MaxPool & (2,1) & (T/4, 64) \\
            Conv (128 kernels) & (4, 1) & (T/4, 128) \\
            Conv (128 kernels) & (4, 1) & (T/4, 128) \\
            MaxPool & (2, 1) & (T/8, 128) \\
            Output (if backbone) & - & (128,) \\
            Linear (if not backbone) & - & (n\_classes,) \\  
            \midrule
            \# Parameters & & $\approx$905k \\
            \bottomrule
        \end{tabular}
    \end{subtable}
\end{table}

\begin{table}[h]
    \setlength{\tabcolsep}{3pt} % Adjust this value to decrease/increase column spacing
    \centering
    \caption{1D U-Net Architecture}
    \begin{tabular}{llll}
        \toprule
        Layer & Kernel Size & Output Size & Input to Layer \\
        \midrule
        Input  & - & (T, 1) & --- \\
        AvgPool1 & (2,1) & (T/2, 1) & Input \\
        AvgPool2 & (4.1) & (T/4, 1) & Input \\
        Conv1D & (3, 1) & (T, 32) & Input \\
        Conv1D$_1$ (down) & (3, 1) & (T/2, 64) & Conv1D \\
        Conv1D$_2$ (down) & (3, 1) & (T/4, 96) & \texttt{concat}[Conv1D$_1$, AvgPool1] \\
        Conv1D$_3$ (down) & (3, 1) & (T/8, 128) & \texttt{concat}[Conv1D$_2$, AvgPool2] \\
        Upsample$_1$ & (2, 1) & (T/4, 128) & Conv1D$_3$ \\
        Conv1D$_4$ (up) & (3, 1) & (T/4, 96) & \texttt{concat}[Upsample$_1$, Conv1D$_2$] \\
        Upsample$_2$ & (2, 1) & (T/2, 96) & Conv1D$_4$ \\
        Conv1D$_5$ (up) & (3, 1) & (T/4, 64) & \texttt{concat}[Upsample$_2$, Conv1D$_1$] \\
        Upsample$_3$ & (2, 1) & (T, 64) & Conv1D$_5$ \\
        Conv1D$_6$ (up) & (3, 1) & (T, 32) & \texttt{concat}[Upsample$_3$, Conv1D] \\
        Out Conv & (3, 1) & (T, 1) & Conv1D$_6$ \\
        \midrule
        \# Parameters & & $\approx$355k \\
        \bottomrule
    \end{tabular}
\end{table}

\clearpage
% -------------------------------------------------- %
% -------------------------------------------------- %
% -------------------------------------------------- %
% -------------------------------------------------- %
% -------------------------------------------------- %
% -------------------------------------------------- %
\subsection{Augmentations}
In this section, we give details about the data augmentations that are applied to the time series tasks.
During our experiments, we searched for the best traditional augmentation technique for a given task.
We searched over common time series augmentation methods in literature (Table~\ref{tab:appendix_augmentations}), and applied them with self-supervised learning baselines.

Specifically, we applied \textit{scaling} for ECG and IMU signals to detect heart rate and step counting, \textit{permutation with noise} for HR detection from PPG signals.
\textit{scaling} and \textit{permutation with noise} for respiratory rate from PPG.
We also observed that while permutation enhances the performance of PPG signals, it significantly diminishes the performance of ECG signals.
\begin{table}[h]
    \renewcommand{\arraystretch}{1.5}
    \centering
    \caption{Common time series augmentations}
    \begin{adjustbox}{width=0.85\columnwidth,center}
    \label{tab:appendix_augmentations}
    \begin{tabular}{ll}
        \toprule
        Augmentation & Details \\
        \midrule
        Noise & Add Gaussian noise sampled from normal distribution, $\mathcal{N}(0, 0.3)$ \\
        Scale & Amplify channels by a random distortion sampled from normal distribution $\mathcal{N}(2, 1.1)$ \\
        Negate & Multiply the value of the signal by a factor of -1 \\
        \multirow{2}{*}{Permute} & Split signals into no more than 5 segments, then permute the segments \\
                                & and combine them into the original shape \\
        \multirow{2}{*}{Resample} & Interpolate the time series to 3 times its original sampling rate \\
                                 & and randomly down-sample to its initial dimensions \\
        Time Flip & Flip the time series in time for all channels, i.e., $\boldsymbol{\mathrm{x}}_{Aug}[n] = \boldsymbol{\mathrm{x}}[-n] $\\
        Random Zero Out & Randomly choose a section to zero out (at most 1/10 of the signal)\\
        Permutation + Noise & Combination of Permutation and Noise with the noted parameters \\ 
        Noise + Scale & Combination of Noise and Scaling\\ 
        \bottomrule
    \end{tabular}
    \end{adjustbox}
\end{table}

SimPer applies interpolation and decimation to change the periodicity of the sample as augmentations.
We limit the speed change---the term authors used for resampling the short and long samples---range to be within [0.5, 2], ensuring the augmented sequence is longer or shorter than a fixed length in the time dimension, following the same implementation in the paper.

\clearpage
% -------------------------------------------------- %
% -------------------------------------------------- %
\section{Additonal Results}
\label{appendix:Additional_Results}

\subsection{BIDMC dataset}
\label{appendix:BIDMC}
The BIDMC dataset exhibits a significant imbalance~\cite{bidmc}, leading self-supervised and supervised learning techniques to essentially leverage the training data statistics, particularly the mean output, during inference.
We, therefore, only report the heuristic and unsupervised learning-based methods in Table~\ref{appendix_tab:performance_resp_bidmc}.
\begin{table}[h]
\caption{Comparison of methods in \textit{Respiration} for BIDMC}
\begin{adjustbox}{width=0.5\columnwidth,center}
\label{appendix_tab:performance_resp_bidmc}
\renewcommand{\arraystretch}{0.4}
\begin{threeparttable}
\begin{tabular}{@{}lllll@{}}
\toprule
\multirow{2}{*}{Method} & \multicolumn{2}{l}{BIDMC (64-second)} & \multicolumn{2}{l}{BIDMC (32-second)} \\ 
\cmidrule(r{7pt}){2-3}  \cmidrule(r{7pt}){4-5}  \\ 
& MAE$\downarrow$ & RMSE$\downarrow$ &  MAE$\downarrow$ & RMSE$\downarrow$ \\
\midrule
\textit{Heuristic} & & & &  \\
Fourier & 4.081 & 4.962 & 5.71 & 7.25 \\
Autocorrelation & 14.05 & 16.23 & 14.36 & 16.64 \\
RobustPeriod & 10.84 & 13.36 & 10.02 & 12.79 \\
\midrule
\textit{No Supervision} & & & & \\
SiNC & 4.11\small$\pm$0.10 & 5.44\small$\pm$0.31 & 6.01\small$\pm$0.98 & 7.33\small$\pm$0.92 \\
Ours & \textbf{3.40}\small$\pm$0.20 & \textbf{4.41}\small$\pm$0.43 & \textbf{3.74}\small$\pm$0.03 & \textbf{4.77}\small$\pm$0.12  \\
\bottomrule
\end{tabular}
  \end{threeparttable}
\end{adjustbox}
\end{table}
As seen in Table~\ref{appendix_tab:performance_resp_bidmc}, our proposed method demonstrates a significant performance improvement over the baselines. Furthermore, the performance gap between our method and another deep learning solution, SiNC, exceeds 25\%.

\subsection{The effect of loss weights}
\label{appendix:loss_weights}
In this section, we investigate the effect of loss weights.
During our experiments, we give the same importance to each regularizer, i.e., the coefficients ($\lambda$, $\nu$) in Equation~\ref{appendix_eq:overall_loss} are set to 1.

\begin{equation}\label{appendix_eq:overall_loss}
    \mathcal{L} =  -\lambda \sum_{f^*}^{} S(\omega) \log S(\omega) + \nu \sum_{f^*}^{} S_X(\omega) \log \frac{S_X(\omega)}{S(\omega)}
    + \sum_{f^{'}}^{} S(\omega),  \hspace{2mm} \omega = 2 \pi f \hspace{1mm} \text{and} \hspace{1mm} f^{'} = \mathbb{U} \setminus f^*
\end{equation}

We observed the effect of two coefficients on the performance by setting them to discrete values of $\{0.5, 1.0, 1.5, 2.0\}$, independently.
The below figures present the results for two different signals, including electrocardiogram and photoplethysmography.
It is evident from these figures that the optimal performance is achieved when the weights for both losses are balanced, rather than assigning greater importance to one over the other.
Moreover, a closer inspection of these figures reveals an important outcome regarding the losses.
When the models are trained while giving more importance to decreasing spectral entropy ($\lambda$) rather than keeping the relative entropy between input and output close ($\nu$), they collapse.
This is noticeable when examining the blue line---increasing the importance of periodicity with the same weight for the other loss---the model error increases significantly.

\setlength{\textfloatsep}{15pt plus 1.0pt minus 10.0pt}
\begin{figure}[h]
  \begin{subfigure}{.31\textwidth}
  \centering
    \includegraphics[width=\linewidth]{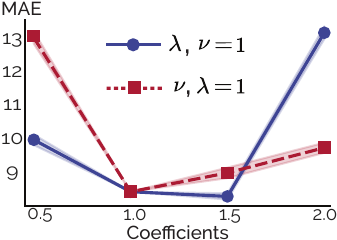}
    \caption{Mean absolute error on IEEE SPC12}
  \end{subfigure}%
  \hspace{1em} % Adjust the space as needed
  \begin{subfigure}{.31\textwidth}
  \centering
    \includegraphics[width=\linewidth]{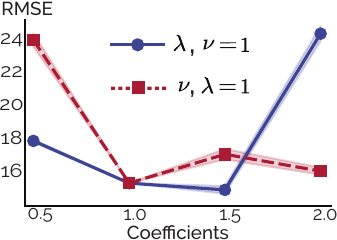}
    \caption{Root mean square error on IEEE SPC12}
  \end{subfigure}
  \hspace{1em} % Adjust the space as needed
  \begin{subfigure}{.31\textwidth}
  \centering
    \includegraphics[width=\linewidth]{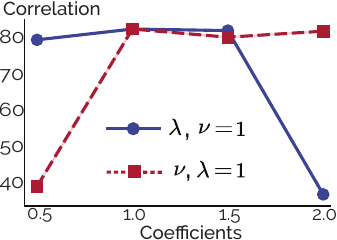}
    \caption{Correlation on IEEE SPC12}
  \end{subfigure}%

  % \begin{subfigure}{.30\textwidth}
  % \centering
  %   \includegraphics[width=\linewidth]{Figures/dataset_KL_resp.pdf}
  %   \caption{Respiratory}
  % \end{subfigure}%  
  % \begin{subfigure}{.33\textwidth}
  % \centering
  %   \includegraphics[width=\linewidth]{Figures/dataset_KL_step.pdf}
  %   \caption{Step counting}
  % \end{subfigure}
    \caption{The three error metrics--—mean absolute error (MAE), root mean square error (RMSE), and correlation—--provided in the SPC12 dataset are used to evaluate the sensitivity of loss weights on the overall performance. }
  \label{fig:loss_sensitivity_SPC12}
\end{figure}

\begin{figure}[h]
  \begin{subfigure}{.31\textwidth}
  \centering
    \includegraphics[width=\linewidth]{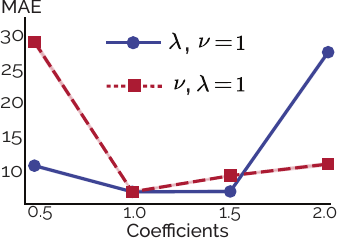}
    \caption{Mean absolute error on IEEE SPC22}
  \end{subfigure}%
  \hspace{1em} % Adjust the space as needed
  \begin{subfigure}{.31\textwidth}
  \centering
    \includegraphics[width=\linewidth]{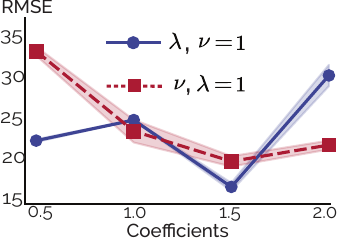}
    \caption{Root mean square error on IEEE SPC22}
  \end{subfigure}
  \hspace{1em} % Adjust the space as needed
  \begin{subfigure}{.31\textwidth}
  \centering
    \includegraphics[width=\linewidth]{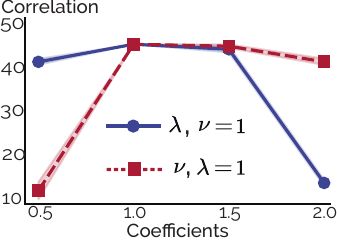}
    \caption{Correlation on IEEE SPC22}
  \end{subfigure}%
  % \end{subfigure}
    \caption{The three error metrics--—mean absolute error (MAE), root mean square error (RMSE), and correlation—--provided in the SPC22 dataset are used to evaluate the sensitivity of loss weights on the overall performance.}
  \label{fig:loss_sensitivity_SPC22}
\end{figure}

Moreover, this effect is more pronounced with the datasets that include more noisy samples.
For example, the mean absolute error increases by twice when more weight is given to periodicity as in Figure~\ref{fig:loss_sensitivity_SPC22}.

\setlength{\textfloatsep}{15pt plus 1.0pt minus 10.0pt}

\setlength{\textfloatsep}{15pt plus 1.0pt minus 10.0pt}
\begin{figure}[h]
  \begin{subfigure}{.31\textwidth}
  \centering
    \includegraphics[width=\linewidth]{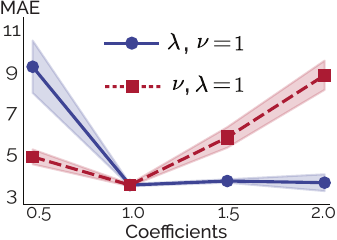}
    \caption{Mean absolute error on PTB-XL}
  \end{subfigure}%
  \hspace{1em} % Adjust the space as needed
  \begin{subfigure}{.31\textwidth}
  \centering
    \includegraphics[width=\linewidth]{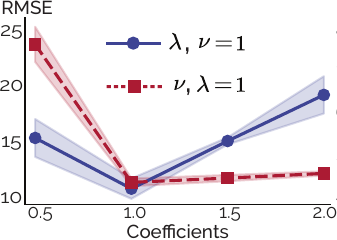}
    \caption{Root mean square error on PTB-XL}
  \end{subfigure}
  \hspace{1em} % Adjust the space as needed
  \begin{subfigure}{.31\textwidth}
  \centering
    \includegraphics[width=\linewidth]{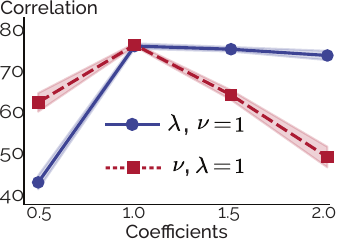}
    \caption{Correlation on PTB-XL}
  \end{subfigure}%
  % \end{subfigure}
    \caption{The three error metrics--—mean absolute error (MAE), root mean square error (RMSE), and correlation—--provided in the PTB-XL dataset are used to evaluate the sensitivity of loss weights on the overall performance.}
  \label{fig:loss_sensitivity_PTB}
\end{figure}

When investigating the model performance with varying loss weights on two signals, we can see the different behavior of the model.
Specifically, in less noisy signal segments such as ECG, increasing the importance of the periodicity loss does not lead to substantial performance declines when compared with PPG signals.
This phenomenon can be attributed to the distinct and clear periodic patterns, such as QRS complexes, present in the ECG signal.
In contrast, the periodicity in blood volume changes is less apparent, hidden beneath higher noise levels, leading the model to capture more noise rather than the desired periodicity when the weights are imbalanced.

\subsection{The effect of architectures}
\label{appendix:supervised_arch}
Here, we present the performances of the baseline architectures that are explicitly given in Appendix~\ref{appendix:Implementation_Details} for each dataset to investigate the effect of model capacity on generalization. 
An interesting result from these tables is that even though the number of parameters increases for architectures, the overall performance decreases for some datasets.
For example, the second version of the fully convolutional neural network has roughly twice the number of parameters when compared with DCL.
However, it performs worse than the DCL in cases such as PTB-XL and DaLiA.
Although we applied this architecture search for the self-supervised learning paradigm as well, we have not observed a performance change across datasets.

\begin{table*}[h]
\centering
\caption{Performance comparison of ours with other methods in \textit{ECG} datasets for HR detection}
\begin{adjustbox}{width=1\columnwidth,center}
\label{appendix_tab:performance_ecg_supervised}
\renewcommand{\arraystretch}{0.7}
\begin{tabular}{@{}lllllllllll@{}}
\toprule
\multirow{2}{*}{Method} & \multicolumn{3}{l}{PTB-XL} & \multicolumn{3}{l}{DaLiA$_{ECG}$} & \multicolumn{3}{l}{WESAD} \\ 
\cmidrule(r{15pt}){2-4}  \cmidrule(r{15pt}){5-7}  \cmidrule(r{15pt}){8-10} \\ 
& MAE$\downarrow$ & RMSE$\downarrow$ & $\rho$$\uparrow$ & MAE$\downarrow$ & RMSE$\downarrow$ & $\rho$$\uparrow$ & MAE$\downarrow$ & RMSE$\downarrow$ & $\rho$$\uparrow$ \\
\midrule
\textit{Supervised} & & & & & & & & & \\
DCL & 6.08\small$\pm$0.78 & 14.23\small$\pm$ 1.43 & 76.40\small$\pm$6.02 & 3.91\small$\pm$0.37 & 6.96\small$\pm$1.06 & 91.42\small$\pm$2.5 & 9.14\small$\pm$0.78 & 7.23\small$\pm$0.75 & 85.44\small$\pm$6.32 \\
DCL$_{v2}$ & 5.65\small$\pm$0.19 & 18.13\small$\pm$ 1.31 & 66.22\small$\pm$0.79 & 2.41\small$\pm$0.06 & 4.91\small$\pm$0.10 & 95.64\small$\pm$0.13 & 2.80\small$\pm$0.09 & 4.49\small$\pm$0.11 & 95.13\small$\pm$0.59 \\
CNN & 9.09\small$\pm$0.27 & 16.37\small$\pm$0.33  & 66.11\small$\pm$1.75 & 5.39$\pm$0.16 & 8.36\small$\pm$0.15 & 89.44\small$\pm$0.72 & 10.69\small$\pm$0.67 & 13.57\small$\pm$1.01 & 50.10\small$\pm$12.5 \\
CNN$_{v2}$ & 7.50\small$\pm$0.45 & 17.36\small$\pm$1.13  & 62.13\small$\pm$5.90 &4.02\small$\pm$0.15 & 6.95\small$\pm$0.37 & 91.87\small$\pm$1.01 & 4.98\small$\pm$0.06 & 7.27\small$\pm$0.19 & 87.98\small$\pm$0.66 \\
U-Net & 11.36\small$\pm$0.50 & 19.33\small$\pm$0.70 & 62.44\small$\pm$2.56 & 4.72\small$\pm$0.03 & 10.88\small$\pm$0.07 & 80.76\small$\pm$0.10 & 5.39\small$\pm$0.02 & 7.75\small$\pm$0.06 & 87.77\small$\pm$0.31 \\
\midrule
\textit{No Supervision} & & & & & & & & & \\
Ours & \textbf{3.75}\small$\pm$0.02 & \textbf{13.58}\small$\pm$0.03 & \textbf{79.30}\small$\pm$0.17 & \textbf{1.29}\small$\pm$0.001 & \textbf{2.08}\small$\pm$0.01 &\textbf{99.12}\small$\pm$0.01& \textbf{2.13}\small$\pm$0.01 & \textbf{3.88}\small$\pm$0.06 & \textbf{93.52}\small$\pm$0.61 & \\
\bottomrule
\end{tabular}
\end{adjustbox}
\end{table*}

\begin{figure}[t]
  \begin{minipage}[h]{0.45\linewidth}
    \centering
    \begin{table}[H]
    \centering
      \caption{Comparison of methods in \textit{Step counting}}
      \begin{adjustbox}{width=1\columnwidth,center}
        \label{appendix_tab:performance_step_supervised}
        \renewcommand{\arraystretch}{0.7}
        \begin{tabular}{@{}lllll@{}}
          \toprule
          \multirow{2}{*}{Method} & \multicolumn{2}{l}{Clemson (Regular)} & \multicolumn{2}{l}{Clemson (Semi-regular)} \\ 
          \cmidrule(r{15pt}){2-3}  \cmidrule(r{5pt}){4-5}  \\ 
          & MAPE$\downarrow$ & MAE$\downarrow$ &  MAPE$\downarrow$ & MAE$\downarrow$ \\
          \midrule
          \textit{Supervised} & & & &  \\
          DCL & 5.99\small$\pm$0.34 & 3.45\small$\pm$0.32 & 19.59\small$\pm$1.54 & 8.98\small$\pm$0.87  \\
          DCL$_{v2}$ & \textbf{5.64}\small$\pm$0.59 & \textbf{3.23}\small$\pm$0.33 & 18.73\small$\pm$0.54 & 7.90\small$\pm$0.35  \\
          FCN & 6.15\small$\pm$0.60 & 3.53\small$\pm$0.33 & 17.22\small$\pm$0.31 & 6.98\small$\pm$0.23 \\
          FCN$_{v2}$ & 5.97\small$\pm$0.26 & 3.48\small$\pm$0.15 & \textbf{15.24}\small$\pm$0.57 & \textbf{6.44}\small$\pm$0.28 \\
          U-Net & 6.24\small$\pm$0.53 & 3.62\small$\pm$0.32 & 16.83\small$\pm$0.45 & 7.71\small$\pm$0.26 \\
          \midrule
          \textit{No Supervision} & & & & \\
          Ours & 5.95\small$\pm$0.21 & 3.42\small$\pm$0.17 & 35.18\small$\pm$0.02 & 14.21\small$\pm$0.97  \\
          \bottomrule
        \end{tabular}
      \end{adjustbox}
    \end{table}
  \end{minipage}
  \hspace{0.5cm} % Adjust the horizontal space between tables
  \begin{minipage}[h]{0.45\linewidth}
    \centering
    \begin{table}[H]
      \caption{Comparison of methods in \textit{Respiration}}
      \begin{adjustbox}{width=1\columnwidth,center}
        \label{appendix_tab:performance_resp_supervised}
        \renewcommand{\arraystretch}{0.63}
        \begin{tabular}{@{}lllll@{}}
          \toprule
          \multirow{2}{*}{Method} & \multicolumn{2}{l}{CapnoBase (64-second)} & \multicolumn{2}{l}{CapnoBase (32-second)} \\ 
          \cmidrule(r{7pt}){2-3}  \cmidrule(r{7pt}){4-5}  \\ 
          & MAE$\downarrow$ & RMSE$\downarrow$ &  MAE$\downarrow$ & RMSE$\downarrow$ \\
          \midrule
          \textit{Supervised} & & & &  \\
          DCL & 5.76\small$\pm$0.28 & 7.45\small$\pm$0.27 & 5.74\small$\pm$0.08 & 7.68\small$\pm$0.07 \\
          DCL$_{v2}$ & 6.28\small$\pm$0.37 & 8.61\small$\pm$0.71 & 6.30\small$\pm$0.80 & 8.71\small$\pm$1.04 \\
          FCN & 6.00\small$\pm$0.20 & 8.15\small$\pm$0.28 & 5.41\small$\pm$0.24 & 7.57\small$\pm$0.40 \\
          FCN$_{v2}$ & 5.49\small$\pm$0.21 & 7.65\small$\pm$0.38 & 5.63\small$\pm$0.60 & 7.73\small$\pm$0.70 \\
          U-Net & 5.49\small$\pm$0.21 & 7.65\small$\pm$0.38 & 5.63\small$\pm$0.60 & 7.73\small$\pm$0.70 \\
          \midrule
          \textit{No Supervision} & & & & \\
          Ours & \textbf{3.40}\small$\pm$0.20 & \textbf{4.41}\small$\pm$0.43 & \textbf{3.74}\small$\pm$0.03 & \textbf{4.77}\small$\pm$0.12  \\
          \bottomrule
        \end{tabular}
      \end{adjustbox}
    \end{table}
  \end{minipage}
\end{figure}

\begin{table*}[t]
\centering
\caption{Performance comparison of supervised methods with different architectures in \textit{PPG} datasets for HR estimation}
\begin{adjustbox}{width=1\columnwidth,center}
\label{appendix_tab:performance_ppg_supervised}
\renewcommand{\arraystretch}{0.9}
\begin{tabular}{@{}lllllllllll@{}}
\toprule
\multirow{2}{*}{Method} & \multicolumn{3}{l}{IEEE SPC12} & \multicolumn{3}{l}{IEEE SPC22} & \multicolumn{3}{l}{DaLiA$_{PPG}$} \\ 
\cmidrule(r{15pt}){2-4}  \cmidrule(r{15pt}){5-7}  \cmidrule(r{15pt}){8-10} \\ 
& MAE$\downarrow$ & RMSE$\downarrow$ & $\rho$$\uparrow$ & MAE$\downarrow$ & RMSE$\downarrow$ & $\rho$ $\uparrow$ & MAE$\downarrow$ & RMSE$\downarrow$ & $\rho$$\uparrow$ \\
\midrule
\textit{Supervised} & & & & & & & & & \\
DCL & 19.90\small$\pm$1.10 & 26.34\small$\pm$1.10 & 25.53\small$\pm$4.2 & 22.43\small$\pm$0.62 & 27.17\small$\pm$0.60 & 11.95\small$\pm$5.1 & 5.97\small$\pm$0.19& 11.83\small$\pm$0.36 & 80.41\small$\pm$0.9\\
DCL$_{v2}$ & 11.05\small$\pm$0.08 & 14.64\small$\pm$0.10 & 77.40\small$\pm$0.71 & 18.89\small$\pm$0.44 & 22.89\small$\pm$0.26 & 09.90\small$\pm$5.1 & \textbf{5.66}\small$\pm$0.06 & \textbf{11.29}\small$\pm$0.21 & \textbf{81.99}\small$\pm$0.46 \\
CNN & 12.48\small$\pm$0.23 & 18.19\small$\pm$0.26 & 68.08\small$\pm$1.85 & 17.97\small$\pm$0.33 & 23.06\small$\pm$0.36 & 21.91\small$\pm$1.76 & 7.35\small$\pm$0.14 & 13.74\small$\pm$0.26 & 74.22\small$\pm$0.55 \\
CNN$_{v2}$ & 12.10\small$\pm$0.22 & 18.19\small$\pm$0.26 & 68.08\small$\pm$1.85 & 18.12\small$\pm$0.37 & 22.90\small$\pm$0.39 & 19.83\small$\pm$3.14 & 6.89\small$\pm$0.02 & 13.02\small$\pm$0.12 & 76.60\small$\pm$0.31 \\
U-Net & 18.40\small$\pm$0.79 & 25.06\small$\pm$1.05 & 39.53\small$\pm$4.76 & 25.86\small$\pm$0.75 & 31.45\small$\pm$0.83 & 06.74\small$\pm$1.76 & 8.70\small$\pm$0.21 & 16.76\small$\pm$0.34 & 65.61\small$\pm$1.11 \\
\midrule
\textit{No Supervision} & & & & & & & & & \\
Ours & \textbf{9.30}\small$\pm$0.10 & \textbf{16.50}\small$\pm$0.20 & \textbf{77.60}\small$\pm$0.43 & \textbf{10.27}\small$\pm$0.37 & \textbf{19.62}\small$\pm$0.71 & \textbf{44.10}\small$\pm$0.89 & 27.41\small$\pm$4.73 & 31.26\small$\pm$4.55 & 18.12\small$\pm$3.86 & \\
\bottomrule
\end{tabular}
\end{adjustbox}
\end{table*}

\subsection{Comparison with additional traditional methods}
\label{appendix:more_traditional}
Here, we compared our approach with a traditional wavelet-based method~\cite{wavelet_based_delineator} specifically designed for detecting the QRS complex in ECG signals, as QRS complexes represent the main source of periodicity in the ECG~\cite{Pan_Tompkins}, which aligns with the periodicity of interest in our experiments for the ECG signals.
The comparison results are given in Table~\ref{appendix:tab_performance_ecg_traditionals}.
\begin{table*}[h]
\centering
\caption{Performance comparison of ours with other methods in \textit{ECG} datasets for HR detection}
\begin{adjustbox}{width=1\columnwidth,center}
\label{appendix:tab_performance_ecg_traditionals}
\renewcommand{\arraystretch}{0.7}
\begin{tabular}{@{}lllllllllll@{}}
\toprule
\multirow{2}{*}{Method} & \multicolumn{3}{l}{PTB-XL} & \multicolumn{3}{l}{DaLiA$_{ECG}$} & \multicolumn{3}{l}{WESAD} \\ 
\cmidrule(r{15pt}){2-4}  \cmidrule(r{15pt}){5-7}  \cmidrule(r{15pt}){8-10} \\ 
& MAE$\downarrow$ & RMSE$\downarrow$ & $\rho$$\uparrow$ & MAE$\downarrow$ & RMSE$\downarrow$ & $\rho$$\uparrow$ & MAE$\downarrow$ & RMSE$\downarrow$ & $\rho$$\uparrow$ \\
\midrule
\textit{Heuristic} & & & & & & & & & \\
Fourier & 10.51 & 28.65 & 49.12 & 4.06 & 15.46 & 36.86 & 4.88 & 18.57 & 56.69\\
Autocorrelation & 8.64 & 19.93 & 59.63 & 7.17 & 14.07 & 38.17 & 4.19 & 8.046 & 89.90 & \\
RobustPeriod & 72.79 & 80.87 & -2.93 & 12.79 & 23.21 & 37.49 & 4.19 & 7.130 & 62.69 & \\
Wavelet-based & 10.06 & 22.85 & 50.73 & 5.41 & 10.94 & 90.52 & 9.47 & 23.19 & 70.45 & \\
\midrule
Ours & \textbf{3.75}\small$\pm$0.02 & \textbf{13.58}\small$\pm$0.03 & \textbf{79.30}\small$\pm$0.17 & \textbf{1.29}\small$\pm$0.001 & \textbf{2.08}\small$\pm$0.01 &\textbf{99.12}\small$\pm$0.01& \textbf{2.13}\small$\pm$0.01 & \textbf{3.88}\small$\pm$0.06 & \textbf{93.52}\small$\pm$0.61 & \\
\bottomrule
\end{tabular}
\end{adjustbox}
\end{table*}

As shown in Table~\ref{appendix:tab_performance_ecg_traditionals}, our proposed method outperforms traditional methods by a significant margin.
we believe that the main reason for this performance difference is the fact that these methods are often developed by optimizing numerous hyperparameters specific to the dataset statistics, making them less effective in a general case.

\clearpage
\section{Visual Examples}
\label{appendix:figures_representations}
Here, we give some examples where the heuristic rule-based methods fail to detect periodicities in signals due to various types of noise.
First, we show the cases when our method performs better than the Pan-Tompkins~\cite{Pan_Tompkins} algorithm for periodicity detection in ECG signals.

\setlength{\textfloatsep}{15pt plus 1.0pt minus 4.0pt}
\begin{figure}[h]
  \begin{subfigure}{\textwidth} 
    \centering
    \includegraphics[width=1\linewidth]{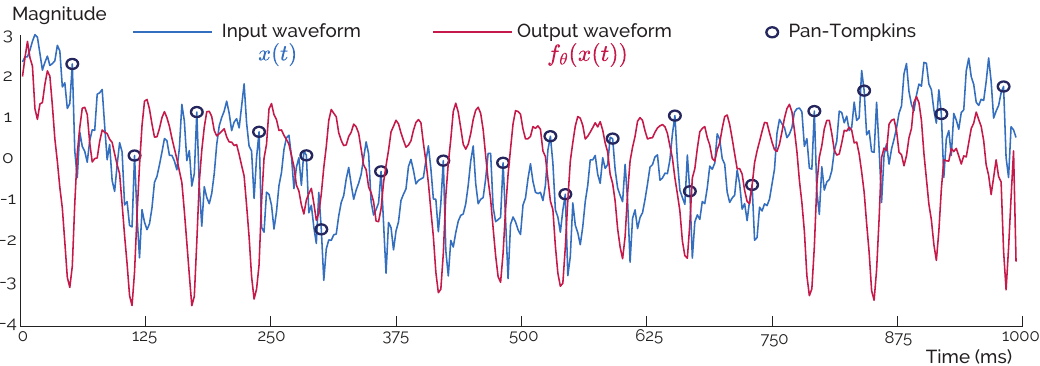} 
    \caption{A 10-second ECG recording from PTB-XL, noise is present with numerous spikes at the same amplitude as the QRS complex, leading to the Pan-Tompkins algorithm detecting multiple beats at the same location.}
    \label{appendix_fig:ECG1_out_inp}
  \end{subfigure}
  
  \vspace{\baselineskip} % Add some vertical space between the subfigures

  \begin{subfigure}{\textwidth} % Use full text width for the second subfigure
    \centering
    \includegraphics[width=1\linewidth]{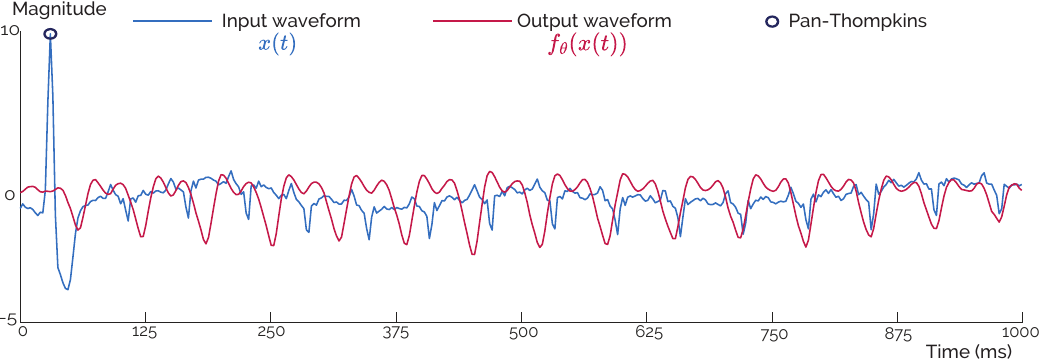} % Adjust the width as needed
    \caption{A recording from PTB-XL that contains an abrupt impulse noise at the beginning, which causes Pan-Tompkins to fail.}
    \label{appendix_fig:ECG2_out_inp}
  \end{subfigure}

  \caption{The visual examples for ECG signals with the raw signal $\mathbf{x}(t)$, the output of the model $f_{\theta}(\mathbf{x}(t))$, and the detected points using Pan-Tompkins algorithm.
  Despite its common use for identifying periodic patterns, QRS complexes, in ECG signals, we present two examples where the Pan-Tompkins algorithm fails due to various types of noise in the recordings.}
  \label{appendix_fig:ECG1_out}
\end{figure}

One interesting outcome of these illustrations is that the model demonstrates the ability to entirely disregard the impulse noise at the beginning (See Figure~\ref{appendix_fig:ECG1_out_inp}), and effectively approximate the periodicity in the signal. 
Moreover, a closer inspection of these two figures shows that the proposed method also works as an unsupervised segmentation algorithm of signals. 
For example, the end of Figure~\ref{appendix_fig:ECG1_out} demonstrates that the output waveform matches the desired patterns.
However, when the time series becomes non-stationary in the segment, the segmentation shifts as the proposed algorithm assumes the signals are periodic or quasi-periodic in the investigated segment.
This assumption is widely used in algorithm development for the human-generated data~\cite{demirel2023chaos}, such as signals generated from the cardiovascular (ECGs, PPGs) or skeletal system (IMUs) of humans similar to our case.
Therefore, we believe that the proposed algorithm can be applied to segmentation tasks, enabling the identification of the desired periodic events in a completely unsupervised manner.

\clearpage

\begin{figure}[h]
  \begin{subfigure}{\textwidth} 
    \centering
    \includegraphics[width=1\linewidth]{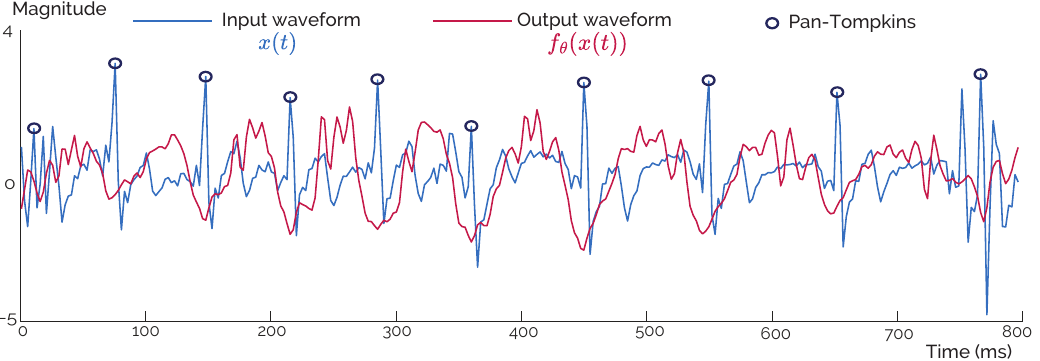} 
    \caption{An 8-second ECG recording from DaLiA $\mathbf{x}(t)$, and the output waveform from the trained model $f_{\theta}(\mathbf{x}(t))$. 
    It can be seen that the recording is noisy at the beginning and the end where the Pan-Tompkins algorithm confuses true periodic patterns with noise in the signal while giving a good approximation of the heart rate. }
    \label{appendix_fig:ECG2_inp}
  \end{subfigure}
  
  \vspace{\baselineskip} % Add some vertical space between the subfigures

  \begin{subfigure}{\textwidth} % Use full text width for the second subfigure
    \centering
    \includegraphics[width=1\linewidth]{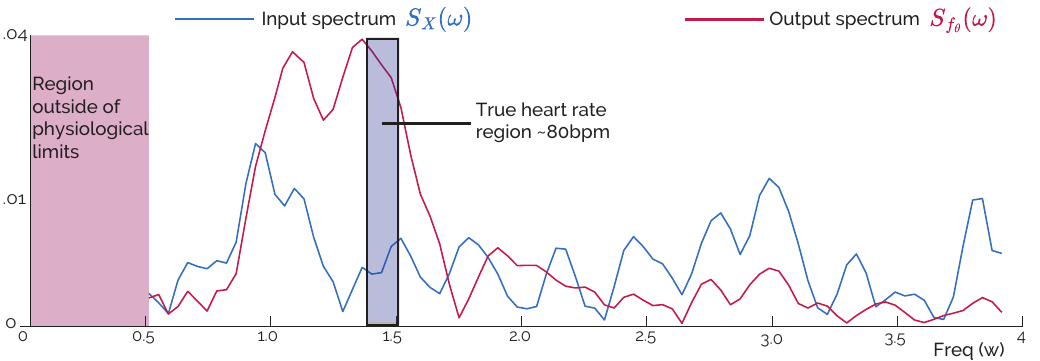} % Adjust the width as needed
    \caption{The spectra of the input and output signal for comparison.
    The frequencies lower than 0.5 are shown with red as they correspond to the heart rate outside of physiological limits, i.e., lower than 30 beats per minute.
    The input spectrum shows the main periodic signal is at around 1\,Hz due to corrupted waveform, especially at the beginning and end.
    A rule-based algorithm depending on the input spectrum may incorrectly identify the desired periodicity due to inaccuracies in the peak location.
    However, the spectrum of the output waveform increased the magnitude of the sinusoidal at the desired periodicity region significantly, close to a 7\,dB increase.
    More importantly, the output waveform includes the desired pattern as the dominant peak rather than noisy ones, which allows the detection of targeted periodicity in the signal.}
    \label{appendix_fig:dalia_wave}
  \end{subfigure}

  \caption{The visual examples for ECG signals with the raw signal $\mathbf{x}(t)$, the output of the model $f_{\theta}(\mathbf{x}(t))$, with their corresponding spectral densities.
  The figures show how the trained model can select and amplify the desired pattern even though it is under the noise level.}
  \label{appendix_fig:dalia_spectra}
\end{figure}

Here, we provided an example of how the proposed method is similar to an adaptive sinusoidal selector based on the Fourier transform.
In uniformly sampled time sequences, periodicity is typically identified by the highest value in
the spectral density~\cite{periodicity_fft}. 
Nevertheless, this method proves ineffective in situations where multiple periodicities exist within the frequency range of interest or noise, as illustrated in Figure~\ref{appendix_fig:dalia_spectra}.

\setlength{\textfloatsep}{15pt plus 1.0pt minus 4.0pt}

According to these two figures, we can infer that the model learns the periodic representations of interest from the time domain waveform such that when it is evaluated with a signal that the model has never seen before, it can detect these patterns and surpass the noise.
The task becomes relatively simpler for ECG signals due to their distinct periodicity indicator, the QRS complex.
This allows the model to learn and effectively ignore impulse noise or other periodic noise sources even though they are both within the frequency range of interest.

\clearpage

Here, we present a more challenging example involving signals generated by blood volume changes.
These signals are susceptible to noise and exhibit less evident periodicity, lacking specific complexes like ECG signals.

\setlength{\textfloatsep}{15pt plus 1.0pt minus 4.0pt}
\begin{figure}[h]
  \begin{subfigure}{\textwidth} 
    \centering
    \includegraphics[width=1\linewidth]{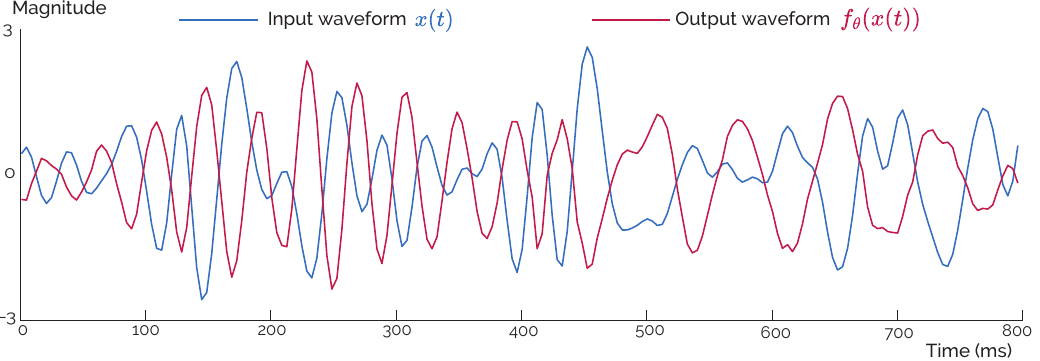} 
    \caption{A 8-second PPG recording from IEEE SPC $\mathbf{x}(t)$, and the output waveform from the trained model $f_{\theta}(\mathbf{x}(t))$. 
    The input waveform is so noisy that the true volumetric blood changes are not observable at all.
    It can be seen from the below figure that the desired periodicity is under the noise level.}
    \label{appendix_fig:PPG_waveform}
  \end{subfigure}
  
  \vspace{\baselineskip} % Add some vertical space between the subfigures

  \begin{subfigure}{\textwidth} % Use full text width for the second subfigure
    \centering
    \includegraphics[width=1\linewidth]{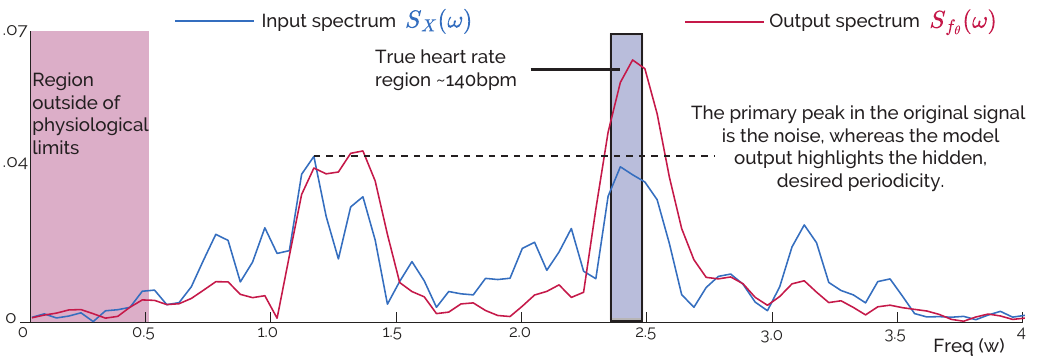} % Adjust the width as needed
    \caption{The spectra of the input and output blood volume changes for comparison.
    The frequencies lower than 0.5\,Hz are shown with red as they correspond to the heart rate outside of physiological limits, i.e., lower than 30 beats per minute.
    Similarly, we limit the highest frequency to 4\,Hz.
    The input spectrum shows the main periodic signal is at around 1.2\,Hz due to the corrupted signal.
    Similar, to the ECG case, a rule-based algorithm depending on the input spectrum may incorrectly identify the desired periodicity due to inaccuracies of the dominant peak in the power spectrum.
    The model outputs a waveform that contains the amplified sinusoidal at the desired periodicity while increasing the overall signal-to-noise ratio.}
    \label{appendix_fig:PPG_fft}
  \end{subfigure}

  \caption{The visual examples for photoplethysmogram signals with the raw waveform $\mathbf{x}(t)$, and the output of the model $f_{\theta}(\mathbf{x}(t))$.
  A waveform example where the Fourier transformation-based periodicity detection fails because of a noisy periodic pattern at the same frequency band.
  In contrast, the proposed model effectively learns and amplifies the pattern of interest for accurate detection.}
  \label{appendix_fig:ppg_spectra}
\end{figure}

What stands out in these two figures is that the model learns the pattern of the desired periodic waveform from the time domain during training such that it can detect these patterns during inference (even though they are under the noise level) and surpass the other periodic noise.

These illustrative examples further support the claim that the proposed method operates as an adaptive sinusoidal selector by learning the periodic source of interest during training and amplifying them during inference without the need for supervision.
We believe the results can be easily improved more by selecting clean samples during training, i.e., curriculum learning, to reinforce the pattern learning, or incorporating physiological limits more effectively into the loss function.

\section{Computational Efficiency}
\label{appendix:efficiency}
In this section, we provide details on the computational efficiency of our proposed method. We employed a U-Net architecture with 300K parameters, making it lightweight and suitable for deployment on various platforms, including mobile devices.
For context, MobileNetV2, often considered a lightweight model in deep learning, has 3.4M parameters, which is approximately 10 times larger than our model.

To provide a comprehensive comparison, we conducted two types of model overhead analyses using the blood volume pulse signals as input similar to the parameter calculations in Appendix~\ref{appendix:arch_details}.
First, we calculated the inference time on a computer equipped with an Intel Core i7-10700k CPU running at 3.80 GHz, 32 GB RAM. 
Second, we deployed our model on a MAX78000 AI Accelerator, which has previously implemented U-Net architectures~\cite{arthur_bell_labs}.
Detailed performance metrics and comparisons with the average of five runs with the standard deviations are provided in Table~\ref{appendix_tab:memory} where the running time is given in milliseconds.

\begin{table}[h!]
    \centering
    \caption{Memory and inference time comparison}
    \label{appendix_tab:memory}
    \begin{tabular}{lcccc}
        \toprule
        \textbf{Method} & \multicolumn{2}{c}{MAX78000} & \multicolumn{2}{c}{PC} \\
        & Memory Requirement & Inference Time & Memory Requirement & Inference Time \\
        \midrule
        FFT & 2.8$\pm$0.1 KB & 0.05$\pm$0.02 & 1.92$\pm$0.2 KB & 0.10$\pm$0.03 \\
        Wavelet & Out-of-Memory & N/A & 10.4$\pm$1.2 MB & 41.1$\pm$3.6 \\
        Robust Period & Out-of-Memory & N/A & 24.7$\pm$3.6 MB & 675$\pm$30.2 \\
        Ours & 19.1$\pm$0.7 KB & 26.9$\pm$0.23 & 16.4$\pm$1.4 KB & 3.2$\pm$0.7 \\
        \bottomrule
    \end{tabular}
\end{table}

An important aspect to note is that PyTorch (or MATLAB) efficiently computes the FFT; when analyzed with the profiler, the memory usage is close to that of the input (in-memory computation).
However, for the MAX78000, our implementation, instead of DSP, requires slightly more memory. Overall, as shown by these experiments, our method not only offers better solutions for periodicity detection but also emphasizes practical implementation by having a very low memory footprint with fast inference.
Moreover, some traditional methods do not fit the MCU flash memory due to heavy wavelet computations.

\end{document}